\pdfoutput=1

\documentclass[11pt]{article}

\usepackage[preprint]{acl}

\usepackage{times}
\usepackage{latexsym}

\usepackage[T1]{fontenc}

\usepackage[utf8]{inputenc}

\usepackage{microtype}

\usepackage{inconsolata}

\usepackage{graphicx}
\usepackage{booktabs}
\usepackage{xcolor}
\usepackage[table]{xcolor}
\usepackage{amsmath} 
\usepackage{amsfonts}
\usepackage{algorithm}
\usepackage{algorithmic}
\definecolor{oodgreen}{RGB}{220, 255, 220}
\definecolor{idred}{RGB}{255, 230, 230}
\usepackage[utf8]{inputenc}
\usepackage[most]{tcolorbox}
\usepackage{xcolor}
\usepackage{listings}
\usepackage{subcaption}
\usepackage{enumitem}
\usepackage{marvosym}
\usepackage{array}
\title{NaviMaster: Learning a Unified Policy for GUI and Embodied Navigation Tasks}

\author{
 \textbf{Zhihao Luo\textsuperscript{1,2}},
 \textbf{Wentao Yan\textsuperscript{1}},
 \textbf{Jingyu Gong\textsuperscript{1}},
 \textbf{Min Wang\textsuperscript{3}},
\\
 \textbf{Zhizhong Zhang\textsuperscript{1}},
 \textbf{Xuhong Wang\textsuperscript{2}}\textrm{\Letter},
 \textbf{Yuan Xie\textsuperscript{1}},
 \textbf{Xin Tan \textsuperscript{1,2}}\textrm{\Letter}
\\
 \textsuperscript{1}East China Normal University,
 \textsuperscript{2}Shanghai AI Laboratory,
 \textsuperscript{3}SenseTime Research
\\
\small{
   {luozhihao@stu.ecnu.edu.cn, wangxuhong@pjlab.org.cn, xtan@cs.ecnu.edu.cn}
 }
}

\begin{document}
\maketitle
\begingroup
\renewcommand{\thefootnote}{\fnsymbol{footnote}}  
\footnotetext[1]{\textrm{\Letter} Corresponding authors: wangxuhong@pjlab.org.cn, xtan@cs.ecnu.edu.cn}
\endgroup
\begin{abstract}
Recent advances in Graphical User Interface (GUI) and embodied navigation have driven progress, yet these domains have largely evolved in isolation, with disparate datasets and training paradigms. In this paper, we observe that both tasks can be formulated as Markov Decision Processes (MDP), suggesting a foundational principle for their unification. Hence, we present NaviMaster, the first unified agent capable of unifying GUI navigation and embodied navigation within a single framework. Specifically, NaviMaster (i) proposes a visual-target trajectory collection pipeline that generates trajectories for both GUI and embodied tasks using a single formulation. (ii) employs a unified reinforcement learning framework on the mix data to improve generalization. (iii) designs a novel distance-aware reward to ensure efficient learning from the trajectories. Through extensive experiments on out-of-domain benchmarks, NaviMaster is shown to outperform state-of-the-art agents in GUI navigation, spatial affordance prediction, and embodied navigation. Ablation studies further demonstrate the efficacy of our unified training strategy, data mixing strategy, and reward design. Our codes, data, and checkpoints are available at \url{https://iron-boyy.github.io/navimaster-page/}.

\end{abstract}

\section{Introduction}
\begin{figure}[t]
   \includegraphics[width=1\columnwidth]{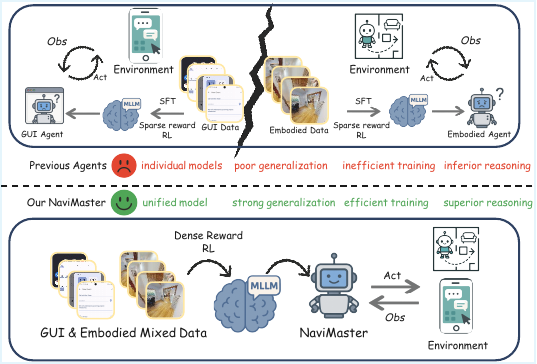}
    \vspace{-8mm}
    \caption{Previous methods involve individual models for GUI and embodied navigation.
   Our NaviMaster is a unified learning framework.}
   \vspace{-4mm}
   \label{fig1}
\end{figure}
Graphical user interface (GUI) navigation agents and embodied navigation agents are designed to traverse virtual and physical environments, respectively. Recent advances in multimodal large language models (MLLMs) \cite{bai2025qwen25vltechnicalreport,Jin_2025} have enabled the integration of their strong perception and planning abilities for both agents \cite{wu2025osatlas,lin2025evolvenavselfimprovingembodiedreasoning}. Leveraging these capabilities, such agents have shown substantial potential in instruction-guided multimodal navigation tasks.

Despite the progress of previous agents, as illustrated in Fig.~\ref{fig1}, the long‑term separation between GUI and embodied navigation, coupled with their training strategies, has resulted in four persistent challenges. (1) These approaches employ two individual models for navigation, which increases training and deployment costs and precludes synergistic interaction between the two tasks \cite{hong2025embodiedwebagentsbridging}. (2) Although prior works~\cite{liu2025idmr,rawles2023androidinthewild,ramakrishnan2021hm3d} have improved performance in respective tasks by scaling data within specific task data, they exhibit limited cross-task performance due to poor generalization to out-of-domain (OOD) data. (3) They face a training‑efficiency bottleneck: previous RFT‑based models employ a sparse reward signal, rendering reinforcement learning optimization inefficient. (4) Current RFT reasoning models often generate correct thoughts but wrong actions, as their “understanding” is primarily distilled from texts rather than visual observations.

To address these challenges, we propose a unified policy that integrates GUI and embodied navigation with an efficient training strategy. Inspired by VIS-Bench \cite{yang2025thinkingspacemultimodallarge}, humans convert egocentric perceptions into an allocentric mental map to support perspective-taking and spatial reasoning. Similarly, both GUI and embodied navigation operate purely on egocentric visual observations, without direct access to a global state. Despite differences in surface-level action semantics, the two tasks are therefore isomorphic at the level of perception and decision-making: in both cases, the agent must integrate partial, first-person observations over time to implicitly construct a latent representation. We therefore characterize the unified problem as learning an egocentric-to-allocentric cognitive transformation. On the other hand, from the perspective of Markov Decision Processes (MDPs):
\begin{equation}
    \arg\max_{a \in \mathcal{A}} P(\mathcal{S}_{t+1} \mid \mathcal{S}_t = \sigma, \mathcal{A}_t = a),
\end{equation}
where next state $\mathcal{S}_{t+1}$ depends solely on the current state–action pair $(\sigma, a)$. Under our unified formulation, the state $\mathcal{S}_t$ corresponds to the egocentric visual observation at step $t$, while the action space $\mathcal{A}$ encompasses interactions in either virtual interfaces or physical environments. The transition dynamics satisfy the Markov property in both domains. Consequently, we formalize GUI and embodied navigation as a single \emph{Navigation Agent} problem.

To this end, as shown in Fig.~\ref{fig1}, NaviMaster incorporates three key advancements (1) We propose a trajectory collection pipeline that unifies both GUI and embodied navigation within a visual-target paradigm, enabling joint training on mix data and improving generalization. (2) We build a unified reinforcement learning training framework applicable to both navigation types. Policies optimized on a single MDP tend to overfit to task-specific correlations. By unifying GUI and embodied navigation under a distribution over MDPs, we enable the policy to learn generalizable structural representations such as visual object permanence, relative spatial reasoning, and affordance grounding. Specifically, we extend the training strategy to estimate task-specific advantages, enabling a single policy to adapt effectively across multiple tasks. Furthermore, the framework leverages prior reasoning steps and actions as historical context. Given the history and current observations as inputs, the model predicts the next action. This formulation unifies the I/O representation and enables the history to guide precise high-level actions in long-horizon navigation. (3) We employ a distance‑aware {dense} reward in the reinforcement learning framework, which enhances training efficiency compared to a sparse binary reward.

We evaluate NaviMaster on OOD GUI and embodied navigation benchmarks. On test sets that differ significantly from the training domain, NaviMaster outperforms state-of-the-art baselines, achieving superior results across multiple datasets. These findings demonstrate strong generalization capability and robustness to distributional shifts. 
In summary, our key contributions are as follows:
\begin{enumerate}[noitemsep]
    \item We propose NaviMaster, the first unified navigation agent that jointly handles both GUI and embodied navigation within one framework.
    \item We develop a visual-target trajectory collection pipeline that aggregates high-quality trajectories from both GUI and embodied navigation, thereby increasing data diversity and enhancing model generalization capability.
    \item We design a distance-aware dense reward and a unified reinforcement learning pipeline, which together enhance data efficiency and further strengthen model grounding ability.
\end{enumerate}



\section{Related Work}
\begin{figure*}[t]
\centering
\includegraphics[width=1.0\textwidth]{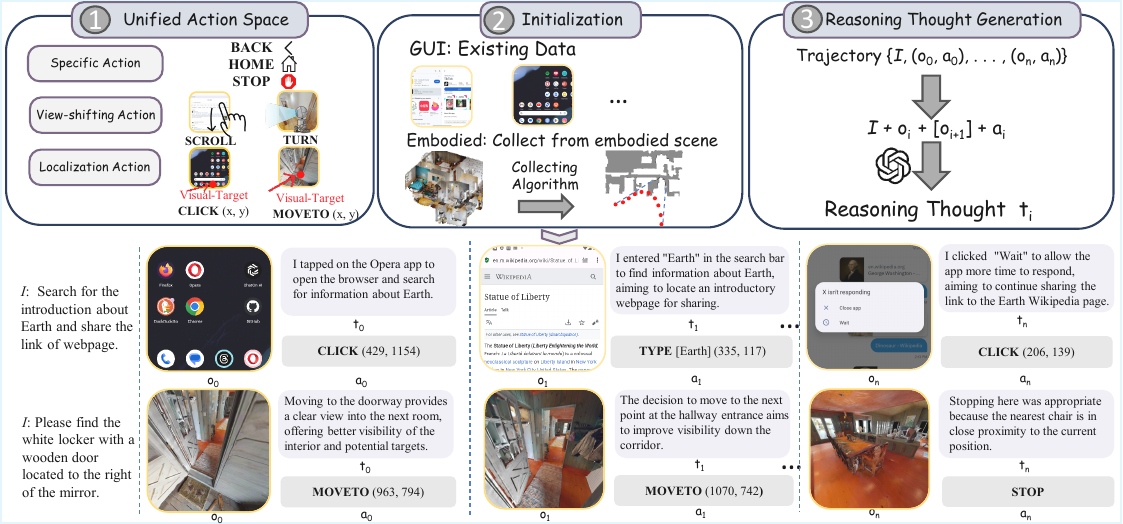} 
\vspace{-8mm}
\caption{Visual-Target Trajectory Collection contains three parts. First, we unify the GUI and the Embodied action space by introducing a visual target at each step. Next, we initialize the trajectories using existing datasets or scenes. Last, we generate a first-person reasoning thought $t_i$ with GPT-4o. Finally, we get our visual-target trajectories $\tau$.}
\label{Dataset Construction}
\vspace{-4mm}
\end{figure*}
\subsection{Navigation Agent}
\textbf{GUI} navigation agents aim to autonomously operate applications by perceiving UI elements and issuing precise point-level actions~\cite{wang2025guiagentsfoundationmodels}. Recent efforts have adopted the data-driven training paradigm such as OS-Atlas, UI-Tars \cite{wu2025osatlas,qin2025uitarspioneeringautomatedgui}. They employ large‑scale datasets in a multi‑stage training pipeline to further enhance their UI grounding precision and planning capability. Despite this progress, most existing GUI agents rely heavily on supervised fine-tuning (SFT) with large amounts of human-annotated data, limiting their generalization capacities. To address this, models like UI-R1 and GUI-R1 \cite{luo2025guir1generalistr1style,lu2025uir1enhancingefficientaction} incorporate reinforcement learning (RL) inspired by DeepSeek-R1. However, their scope remains restricted to GUI-only settings and lack the capacity to do embodied navigation.

\textbf{Embodied} navigation agents control physical or simulated agents to follow language instructions in 3D spaces~\cite{zhang2026worldminecraft,ji2026s2dsparsedenselifting,tian2025drivingforward,ji2025fastlgs}, requiring multimodal perception and long‑horizon planning \cite{gao2024visionlanguagenavigationembodiedintelligence,li2026compassnav,wang2026explorelongtermmemorybenchmark}. Analogous to GUI navigation tasks, works on embodied navigation typically employ a multi-stage SFT strategy on large datasets to adapt open-source MLLMs for navigation tasks (e.g., RoboPoint \cite{yuan2024robopointvisionlanguagemodelspatial}, SpaceLLaVa \cite{foutter2025spacellavavisionlanguagemodeladapted}). Their scope remains restricted to a single domain, which enforces a monolithic action space, thereby limiting the agents' capacity to generalize when the action space changes. 

Recently, Embodied Web Agent (EWA) \cite{hong2025embodiedwebagentsbridging} is the first work that unifies physical embodiment with live web interfaces. Although EWA unifies web and embodied tasks, it lacks an emphasis on grounding capabilities and fails to establish a comparable action space between the two navigation agent types. It also relies on zero/few-shot MLLMs without a unified navigation training paradigm, limiting its value for developing general-purpose navigation agents. 

\subsection{Reinforcement Fine-Tuning on MLLM}
Visual-RFT \cite{liu2025visualrftvisualreinforcementfinetuning,gao2026tpru} performs reinforcement fine-tuning on LVLMs using their own reasoning traces together with rule-based, verifiable visual rewards~\cite{gan2026androidcoachimproveonline,chen2025guishepherdreliableprocessreward}. UI-R1 \cite{lu2025uir1enhancingefficientaction} introduces a unified, rule-based reward that measures the click-coordinate accuracy within the ground-truth bounding box, thereby enhancing the precision of GUI action prediction. GUI-R1 \cite{luo2025guir1generalistr1style} also adopts a similar reward design, but places greater emphasis on high-level GUI navigation capabilities. However, their reward design is strictly binary; only responses that fall within the ground truth bounding box receive a positive score. This leads to many rollouts in GRPO yielding zero reward, making the training process less effective \cite{zheng2025actpaysefficientreinforcement}. Differently, we adopt a dense reward approach for grounding training in navigation agents. Unlike prior work that relies on binary rewards, our method assigns scores based on the proximity of the response to the ground truth, thereby improving grounding performance while promoting more efficient and stable training.

\section{NaviMaster} 

Our proposed NaviMaster consists of three key components, including (1) the visual-target trajectory collection to reformulate the GUI and embodied navigation trajectories into a unified form with historical information, (2) the unified reinforcement learning framework to optimize the cross-scenario at the same time, and (3) the distance-aware reward to update the model parameters by additionally considering the distance between output points and target points.

\subsection{Visual-Target Trajectory Collection}

As Fig.~\ref{Dataset Construction} shows, the visual-target trajectory collection has three parts, including unified action space definition, unified trajectory initialization, and reasoning thought generation. 

\textbf{Unified Action Space Definition.}
Existing GUI and embodied trajectory datasets exhibit substantial differences in their action spaces. We categorize actions into three types: specific action, view-shifting action, and localization action. First, specific actions have predefined, context-independent semantics (e.g., [\textbf{BACK}] in GUI, [\textbf{STOP}] in embodied tasks) and are directly integrated into the unified action space. Second, view-shifting actions adjust the agent’s viewpoint to locate targets outside the current field of view. This class encompasses actions like [\textbf{SCROLL}] in GUI and [\textbf{TURN}] in embodied environments. We standardize these transformations into four directions for each domain: [up, down, left, right] for GUI and [left, right, around, back] for embodied agents. Finally, localization actions show the greatest divergence. In GUI tasks, the localization action is performed through the [\textbf{CLICK} (x, y)] action, where (x, y) denotes a specific target position on the screenshot. In contrast, embodied navigation tasks achieve localization via the [\textbf{MOVEFORWARD}] action, which does not require an explicit target.
These differences reflect distinct interaction paradigms between the two localization actions. GUI actions $\mathcal{A}_{gui}$ depend on precise, target-oriented operations (e.g., clicking specific UI elements), whereas embodied actions  $\mathcal{A}_{emb}$ focus on egocentric motion control (e.g., navigating without explicit target selection).


The discrepancy in localization actions (i.e., with or without a target) creates a challenge for unifying both tasks. To address this challenge, we propose the visual-target trajectory by introducing a localization action with an explicit target into the embodied navigation task. As shown in the left part of Fig.~\ref{Dataset Construction}, we define 
a visual target within the observation at each step of the trajectory. Consequently, the localization action in embodied navigation if reformulated from [\textbf{MOVEFORWARD}] to [\textbf{MOVETO} (x, y)], where (x, y) denotes target location. The complete action space is in Appendix~\ref{action space}.


\textbf{Trajectory Collection Initialization.} 
We consider a long-horizon task consisting of $n$ steps, represented
as $\{I, (o_{0}, a_{0}), \ldots, (o_{n}, a_{n})\}$, where $I$ denotes the user-provided instruction, $o_i$ is the observation from GUI screenshot or physical environment at step $i$, and $a_i$ is the corresponding action at step $i$ ($0\leq i \leq n$). 
This representation is consistent with standard formulations of GUI trajectories, enabling direct reuse of existing GUI datasets. In our experiments, we leverage GUI-Odyssey~\cite{lu2024guiodysseycomprehensivedataset} to obtain trajectory data.

However, the existing embodied navigation datasets typically provide only the initial and the target positions without specifying the intermediate trajectories. For an embodied navigation dataset (e.g.,  Matterport 3D dataset with the Habitat simulator~\cite{yadav2023habitatmatterport3dsemanticsdataset, savva2019habitatplatformembodiedai}), we extract the set of trajectory points along the shortest path from the initial to the target position, denoted as $(\bf{s_{0}}, \bf{s_{1}}, \ldots, \bf{s_{m}})$, which can be mapped by performing the A* search method ~\cite{4082128}. Each trajectory point $\bf{s_{k}}$ ($0 \leq k \leq m$) is a 3D coordinate in the global coordinate system.

Then, based on the point set, we collect observation images and generate visual-target actions for embodied navigation. The initialization procedure for trajectory collection is summarized in Algorithm~\ref{Embodied Trajectory Collection}.
The first step is to align the next position with current observation. Given the current position $\bf{s_k}$$(u_k, v_k, w_k)$ (global coordinate system), its camera rotation $\bf{r_k}$ and the next position $\bf{s_{k+1}}$$(u_{k+1}, v_{k+1}, w_{k+1})$ (global coordinate system), $\bf{s_{k+1}^{'}}$$(u'_{k+1}, v'_{k+1}, w'_{k+1})$ ($\bf{s_k}$ coordinate system) can be obtained with following equation:
\begin{equation}
\bf{s_{k+1}^{'}}=\bf{r_{k}^{-1}}\times(\bf{s_{k+1}}-\bf{s_{k}})\times \bf{r_{k}}.
\end{equation}

After that, we project $\bf{s_{k+1}^{'}}$ onto the current camera observation $o_i$:
\begin{equation}
p(x_i, y_i)=(\frac{W}{2}+f\cdot\dfrac{u'_{k+1}}{w'_{k+1}}, \frac{H}{2}+f\cdot\dfrac{v'_{k+1}}{w'_{k+1}}),
\end{equation}
where $p(x_i, y_i)$ represents the pixel coordinates of the next position in the current observation $o_i$, $(W, H)$ is the width and height of the image, and $f$ is the camera focal length.

Due to the limitations of the camera's pitch angle and field of view, the projected coordinates may not appear within the observation. To address this, we define several custom actions to adjust the camera angle, including the left-right and up-down turning actions in embodied tasks: [\textbf{TURN} left], [\textbf{TURN} right], [\textbf{TURN} around], [\textbf{TURN} down] and implement them as Algorithm \ref{Embodied Trajectory Collection}. Let $w'_{k+1}$ denote the depth of $s'_{k+1}$ in the current observation. A negative value ($w'_{k+1} < 0$) indicates that $s'_{k+1}$ lies behind the camera.
After getting the observation $o_i$ with the target at each position, we represent each embodied navigation trajectory with the same action space and style as in GUI navigation.

\begin{algorithm}[!h]
    \scriptsize
    \caption{Trajectory Collection Initialization for Embodied Task}
    \label{Embodied Trajectory Collection}
    \renewcommand{\algorithmicrequire}{\textbf{Input:}}
    \renewcommand{\algorithmicensure}{\textbf{Output:}}
    
    \begin{algorithmic}[1]
        \REQUIRE \makebox[0.45\linewidth][l]{$I$, $(\bf{s_{0}}, \bf{s_{1}}, \ldots, \bf{s_{m}})$}%
        \ENSURE \makebox[0.45\linewidth][l]{Trajectory}
        
        \STATE Initialize Trajectory with ($I$), $i$ with 0
        
        \FOR{each $k \in [0, m]$}
            \IF {$k < n$}
                \STATE Calculate $\bf{s'_{k+1}}$$(u'_{k+1}, v'_{k+1}, w'_{k+1})$
                \WHILE{$p(x_i,y_i) \notin [0, W] \times [0, H]$}
                    \STATE $o_i \leftarrow$ observation at ($s_k$, $r_k$) 
                    \STATE Update $p(x_i,y_i)$
                    \IF {$w'_{i+1} < 0$} \STATE $a_j \leftarrow$ \textbf{TURN} [around]
                    \ELSIF {$x_i < 0$}\STATE $a_i \leftarrow$ \textbf{TURN} [left]
                    \ELSIF {$x_i > W$}
                        \STATE $a_i \leftarrow$ \textbf{TURN} [right]
                    \ELSIF {$y_i > H$}
                        \STATE $a_i \leftarrow$ \textbf{TURN} [down]
                    \ENDIF
                    \STATE Append ($o_i$, $a_i$) to Trajectory
                    \STATE Update $r_k \leftarrow$ execute $a_i$ ; $i \leftarrow i+1$
                \ENDWHILE
                \STATE $a_i \leftarrow$ \textbf{MOVETO} $(x_i,y_i)$
                \STATE Append ($o_i$, $a_i$) to Trajectory ; $i \leftarrow i+1$
            \ELSE
                \STATE $o_i \leftarrow$ observation at ($s_k$, $r_k$), $a_i \leftarrow$ \textbf{STOP}
                \STATE Append ($o_i$, $a_i$) to Trajectory
            \ENDIF
        \ENDFOR
        
        \RETURN Trajectory
    \end{algorithmic}
\end{algorithm}
\textbf{Reasoning Thought Generation.} 
Historical information has been shown to be beneficial for agent performance~\cite{yang2025embodiedbenchcomprehensivebenchmarkingmultimodal}.  Most existing approaches~\cite{xu2025aguvisunifiedpurevision} take the implemented actions as history, which may lead to ambiguity. For example, the action [\textbf{CLICK} (x, y)] does not specify the context or purpose of the interaction. By contrast, pairing the reasoning thought ``I should first open Chrome to start my search'' with the corresponding action ``[\textbf{CLICK} (x, y)]'' explicitly indicates that agent opens the app Chrome. Some works~\cite{qin2025uitarspioneeringautomatedgui} demonstrates that augmenting each step in the trajectory with its associated reasoning allows the model to articulate its decision-making process more transparently.

To enhance reasoning capability and optimize memory usage, we generate thought for each action in the trajectory, as illustrated in the bottom part of Fig.~\ref{Dataset Construction}.
Given an initialized trajectory, we construct the data generation pipeline as follows:
\begin{equation}
\langle I,o_i,a_i,[o_{i+1}]\rangle\xrightarrow{\mathcal{M}}t_i.
\end{equation}
where the task instruction $I$, observation $o_i$, and action $a_i$ are provided to the large language model $\mathcal{M}$, which produces an intention $t_i$ from a first-person perspective to explain the rationale behind action $a_i$. In our experiments, $\mathcal{M}$ corresponds to GPT-4o~\cite{openai2024gpt4ocard}.  
The optional observation $o_{i+1}$ is included only in GUI trajectories, where the target observation serves as a reference for data generation. The prompts used to generate these reasoning thoughts are provided in Appendix~\ref{Prompts for reasoning thought generation}.

Consequently, the visual-target trajectory for both GUI and embodied navigation tasks is represented as:
 $\tau = \{I, (o_{0}, t_{0}, a_{0}), \ldots, (o_{n}, t_{n}, a_{n})\}$.

\subsection{Unified Reinforcement Learning Framework}



\begin{figure}[t]
\centering
\includegraphics[width=\columnwidth]{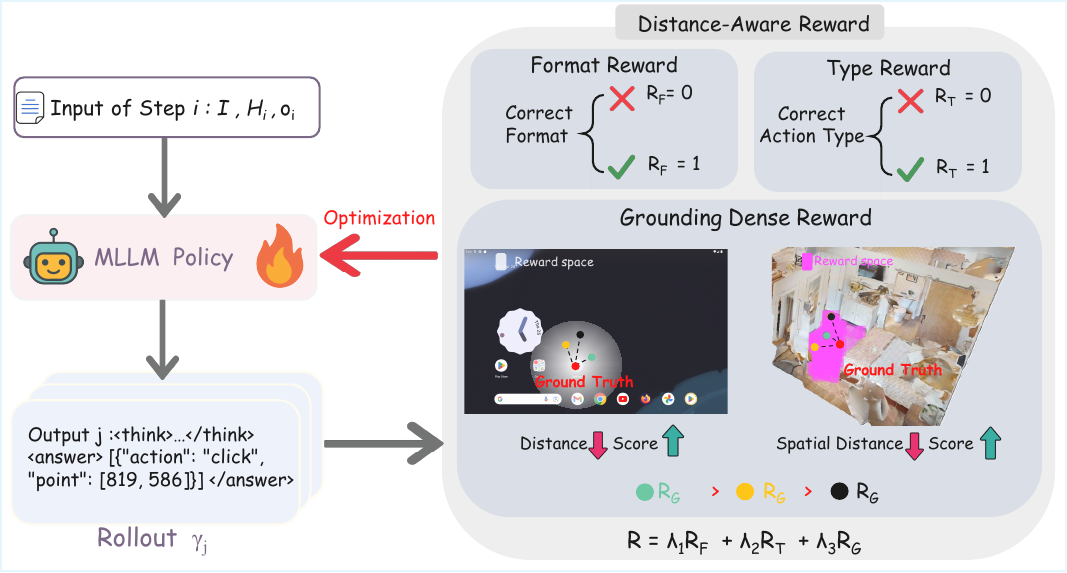} 
\vspace{-8mm}
\caption{{Overview of unified reinforcement learning framework. MLLM policy is optimized using GRPO with format, type and grounding dense reward. }}
\label{mainbody}
\vspace{-5mm}
\end{figure}

Since reinforcement learning (RL) generally exhibits stronger generalization capabilities than supervised fine-tuning (SFT) , we adopt the R1-Zero training strategy, directly training on our collected dataset with Group Relative Policy Optimization (GRPO).
As illustrated in Fig.~\ref{mainbody}, given an n-step trajectory, we take step $i$ as a data sample. Each sample consists of the user instruction $I$, the current observation $o_i$, reasoning thoughts and actions in history $H_i = \{(t_0, a_0), (t_1, a_1), ..., (t_{i-1}, a_{i-1})\}$. NaviMaster then learns a unified policy with GRPO. Specifically, for the input queries $\{I, H_i, o_i\}$, we operate on $G$ samples $\{\gamma_j = \pi_{\theta_{\mathrm{old}}}\left(a_{i}|I, H_i, o_i\right)\}_{j=1}^G$ produced by the policy model $\pi_{\theta}$. We also incorporate the depth map $h_i$ of observation $o_i$ as a critical prior for grounding in spatial space; for pure 2-D images $h_i$ is set to the zero matrix. The advantage $Adv$ is computed as follows:
\begin{align}
&R(i, j) = R(\gamma_j, a_i, h_i),\\
&Adv = \frac{R(i, j) - \text{mean}(\{R(i, j))\}_{j=1}^G)}{\text{std}(\{R(i, j)\}_{j=1}^G)},
\end{align}
where $R(i, j)$ denotes the reward function of the response. It will be detailed in the next section.
\subsection{Distance-Aware Reward}
The criteria for successful task execution are threefold: (1) the model output must be correctly parsed into an executable action, (2) the type of the executed action must match the ground truth, (3) the action’s arguments must fall within valid bounds. Accordingly, we decompose the reward into three components: format, type, and grounding. While most existing reward designs employ binary success/failure signals, our approach aims to capture relative preferences even among unsuccessful rollouts. For instance, among unsuccessful rollouts, some may still be ``better'' than others. Specifically, we design a distance-aware dense reward for the grounding component based on the distance to the ground-truth point. Therefore, our reward consists three components, including format reward, type reward and grounding dense reward. \\
\textbf{Format Reward.} \ This reward $R_F(i, j)$ enforces the formatting of the output. Each response must first provide a reasoning phase, followed by a final answer, where the answer must be a valid JSON string. The required structure is: ``$\langle think \rangle$\allowbreak$  \cdots $\allowbreak$ \langle / think \rangle $\allowbreak$ \langle answer \rangle \ json \ string \ \langle / answer \rangle$''. If a rollout satisfies this format, $R_F(i, j)$ will be set to 1. Otherwise, $R_F(i, j)$ will be set to 0.\\
\textbf{Type Reward.} \ This reward $R_T(i, j) = [\hat{a}_j = a_i]$ evaluates the correctness of the model's action selection. $\hat{a}_j$ denotes the predicted action type for sample $\gamma_j$ and $a_i$ denotes the ground truth action type at step $i$. $[\cdot]$ stands for the Iverson bracket, an indicator that equals 1 when the statement inside is true and 0 otherwise. It is a binary reward that assesses whether the predicted action type matches the ground truth within a small, discrete action space. It supervises the model’s ability to make high-level decisions aligned with task semantics.\\
\textbf{Grounding Dense Reward.} \ This reward $R_G$ is designed to guide model's grounding ability. Specifically, this ability requires selecting the correct target within a large selection space, such as a pixel-level coordinate within an image. It evaluates the predicted location relative to the ground truth at step $i$. To consistently measure grounding performance across navigation tasks, we define a distance-based dense reward instead of a sparse reward. It ensures that the agent receives higher rewards when its prediction is closer to targets (UI elements or position in embodied scene).  Such design can provide effective guidance during training. This reward function is as follows:
\begin{equation}
R_{G}(i, j) = (1-\frac{d_j}{\theta_{d}})[d_j<\theta_{d}, p_j < \theta_{h}]
\end{equation}
where $\theta_d$ and $\theta_h$ are thresholds for distance $d_j$ and depth disparity $p_j$. The $d_j$ denotes the pixel-level distance between predicted point $(\hat{x}_j, \hat{y}_j)$ of $\gamma_j$ and the corresponding ground truth point $(x_i, y_i)$. In embodied environments, the depth value $h_i (\hat{x}_j, \hat{y}_j)$ is also incorporated to account for potential occlusions: two pixels that are close in the 2D image may have substantially different depths in the 3D scene. The details of thresholds are in Appendix~\ref{Discussion of Threshold}. The definitions of $d_j$ and $p_j$ are as follows:
\begin{equation}
d_j = \sqrt{(\hat{x}_j-x_i)^2+(\hat{y}_j-y_i)^2},
\end{equation}
\begin{equation}
p_j = |h_i (\hat{x}_j, \hat{y}_j)-h_i (x_i, y_i)|, 
\end{equation}
The overall reward function is a weighted combination of the three components described above. Here, $R(i, j) = \lambda_{1} R_{F}(i, j) + \lambda_{2} R_{T}(i, j) + \lambda_{3} R_{G}(i, j)$. \( \lambda_{1}, \lambda_{2}, \lambda_{3} \in \mathbb{R}_{+} \) controlling their relative importance of each term.

\section{Experiments}
\subsection{Implementation Details}
We trained our model using the EasyR1 framework~\cite{zheng2025easyr1}, adopting Qwen2.5VL-7B model~\cite{bai2025qwen25vltechnicalreport} as the base model. The training is conducted for three epochs on 8 NVIDIA A800 GPUs, with a global batch size of 128 and a learning rate of $1\times 10^{-6}$. The hyperparameters $\lambda_{1}$, $\lambda_{2}$, and $\lambda_{3}$ are experimentally set to 0.1, 1, and 1, respectively. The training utilizes 20k samples, including 10k GUI samples from GUI-Odyssey and 10k embodied samples from Matterport 3D and RoboPoint. All the experiments of NaviMaster use the same amount of data for a fair comparison. More details are provided in Appendix~\ref{Hyperparameter}. 

\renewcommand{\arraystretch}{0.5}
\begin{table*}[t]
    \centering
    \tiny
    \setlength{\tabcolsep}{2.5pt}
    \setlength{\cmidrulekern}{0.3em}
    \begin{tabular}{l|cccccccccccc|cccc|cc|cc}
        \toprule
        \textbf{Models} & \multicolumn{12}{c|}{\textbf{Mobile}}& \multicolumn{4}{c|}{\textbf{Web}}& \multicolumn{2}{c|}{\textbf{Desktop}}& \multicolumn{2}{c}{\textbf{ID}}\\
        \cmidrule(lr){2-13} \cmidrule(lr){14-17} \cmidrule(lr){18-19} \cmidrule(lr){20-21}
        & \multicolumn{2}{c}{\textbf{AC-Low}} & \multicolumn{2}{c}{\textbf{AC-High}} & \multicolumn{2}{c}{\textbf{AITW}}& \multicolumn{2}{c}{\textbf{GuiAct-P}} & \multicolumn{2}{c}{\textbf{Llamatouch}} & \multicolumn{2}{c|}{\textbf{AITZ}} & \multicolumn{2}{c}{\textbf{GuiAct-W}} & \multicolumn{2}{c|}{\textbf{OmniAct-W}} & \multicolumn{2}{c|}{\textbf{OmniAct-D}}& \multicolumn{2}{c}{\textbf{Odyssey}}\\
        \cmidrule(lr){2-3} \cmidrule(lr){4-5} \cmidrule(lr){6-7} \cmidrule(lr){8-9} \cmidrule(lr){10-11} \cmidrule(lr){12-13} \cmidrule(lr){14-15} \cmidrule(lr){16-17} \cmidrule(lr){18-19} \cmidrule(lr){20-21}
        & \textbf{GR}  & \textbf{SR} & \textbf{GR}  & \textbf{SR} & \textbf{GR}  & \textbf{SR} & \textbf{GR}  & \textbf{SR} & \textbf{GR}  & \textbf{SR} & \textbf{GR}  & \textbf{SR}& \textbf{GR}  & \textbf{SR}& \textbf{GR}  & \textbf{SR}& \textbf{GR}  & \textbf{SR}& \textbf{GR}  & \textbf{SR} \\
        \midrule
        GPT-4o & 38.67  & 28.39 & 30.90 & 21.19 & 35.75  & 26.07 & 26.15  & 26.79 & 46.32 & 38.77 & 27.31  & 20.99 & 45.02 &41.84 & 42.79 &34.06 & 63.25 &50.67 &  14.17 &5.36 \\
        Qwen2.5VL-7B & 87.08  & 62.50 & 59.71 & 47.06 & 65.99  & 45.80 & 55.79  & 44.99 & 75.84 & 60.97 & 60.79  & 33.88 & 90.78  & 78.08 & 84.40  & 71.91 & 79.42  & 57.58 & 68.52  & 37.32 \\
        Qwen2.5VL-7B* & \cellcolor{oodgreen}66.73  & \cellcolor{oodgreen}43.82 & \cellcolor{oodgreen}51.29 & \cellcolor{oodgreen}31.79 & \cellcolor{oodgreen}62.51  & \cellcolor{oodgreen}48.81 & \cellcolor{oodgreen}52.21  & \cellcolor{oodgreen}47.75 & \cellcolor{oodgreen}53.88 & \cellcolor{oodgreen}45.56 & \cellcolor{oodgreen}58.36  & \cellcolor{oodgreen}31.18 & \cellcolor{oodgreen}60.26  & \cellcolor{oodgreen}43.26 & \cellcolor{oodgreen}47.61  & \cellcolor{oodgreen}44.36 & \cellcolor{oodgreen}50.58  & \cellcolor{oodgreen}42.29 & \cellcolor{idred}40.96  & \cellcolor{idred}29.38 \\
        OS-Atlas-7B & \cellcolor{oodgreen}73.37 & \cellcolor{oodgreen}50.94 & \cellcolor{oodgreen}54.90  & \cellcolor{oodgreen}29.83 & \cellcolor{oodgreen}64.89 & \cellcolor{oodgreen}41.38 & \cellcolor{oodgreen}58.52  & \cellcolor{oodgreen}29.43 & \cellcolor{oodgreen}59.25  & \cellcolor{oodgreen}30.11 &\cellcolor{idred}58.30 & \cellcolor{idred}27.58 & \cellcolor{oodgreen}75.61  & \cellcolor{oodgreen}57.02 & \cellcolor{oodgreen}69.35 & \cellcolor{oodgreen}59.15 & \cellcolor{oodgreen}62.87  & \cellcolor{oodgreen}56.73 & \cellcolor{oodgreen}39.74  & \cellcolor{oodgreen}26.96 \\
        Aguvis &\cellcolor{idred}76.56  & \cellcolor{idred}57.55 & \cellcolor{idred}72.09  & \cellcolor{idred}49.96 & \cellcolor{idred}69.57 & \cellcolor{idred}53.06 & \cellcolor{oodgreen}63.39  & \cellcolor{oodgreen}42.66 & \cellcolor{oodgreen}75.08  & \cellcolor{oodgreen}60.41 &\cellcolor{idred}62.61 & \cellcolor{idred}36.15 & \cellcolor{idred}37.18  & \cellcolor{idred}27.66 & \cellcolor{idred}54.27 & \cellcolor{idred}48.29 & \cellcolor{idred}26.00  & \cellcolor{idred}23.31 & -  & - \\
        infiGUI-3B &\cellcolor{idred}93.20&\cellcolor{idred}92.10&\cellcolor{idred}74.40&\cellcolor{idred}71.10&\cellcolor{oodgreen}\textbf{75.58} & \cellcolor{oodgreen}46.51& \cellcolor{oodgreen}\textbf{66.67}& \cellcolor{oodgreen}40.09&\cellcolor{oodgreen}75.02&\cellcolor{oodgreen}58.66&\cellcolor{oodgreen}75.42&\cellcolor{oodgreen}40.81& \cellcolor{oodgreen}86.02  & \cellcolor{oodgreen}64.60 & \cellcolor{oodgreen}75.42  & \cellcolor{oodgreen}54.91 & \cellcolor{oodgreen}73.25  & \cellcolor{oodgreen}47.17 & \cellcolor{oodgreen}70.30  & \cellcolor{oodgreen}33.15 \\
        GUI-R1-7B & \cellcolor{idred}84.02  & \cellcolor{idred}66.52 & \cellcolor{idred}70.31  & \cellcolor{idred}51.56 & \cellcolor{oodgreen}62.94 & \cellcolor{oodgreen}55.31 & \cellcolor{oodgreen}47.09  & \cellcolor{oodgreen}48.62 & \cellcolor{oodgreen}69.82  & \cellcolor{oodgreen}61.27 &\cellcolor{idred}64.98 &\cellcolor{idred}49.06& \cellcolor{oodgreen}88.06 & \cellcolor{oodgreen}74.54  &\cellcolor{oodgreen} 84.87  & \cellcolor{oodgreen}68.69 & \cellcolor{oodgreen}81.19  & \cellcolor{oodgreen}57.70 & \cellcolor{oodgreen}71.10  & \cellcolor{oodgreen}36.22\\
        UI-shift &\cellcolor{idred}94.22 &\cellcolor{idred}73.38 &\cellcolor{idred}73.41 & \cellcolor{idred}52.16&\cellcolor{oodgreen}74.44 &\cellcolor{oodgreen}54.38 &\cellcolor{oodgreen}59.10 &\cellcolor{oodgreen}52.12 &\cellcolor{oodgreen}75.08 &\cellcolor{oodgreen}61.71 &\cellcolor{oodgreen}73.84 & \cellcolor{oodgreen}46.78& \cellcolor{oodgreen}89.49  & \cellcolor{oodgreen}79.43 & \cellcolor{oodgreen}82.93  & \cellcolor{oodgreen}64.22 & \cellcolor{oodgreen}\underline{80.01}  & \cellcolor{oodgreen}57.94 & \cellcolor{oodgreen}62.54  & \cellcolor{oodgreen}32.75 \\
        UI-AGILE &\cellcolor{idred}93.71 &\cellcolor{idred}63.32 &\cellcolor{idred}77.55 & \cellcolor{idred}50.97&\cellcolor{idred}75.72 &\cellcolor{idred}48.66 &\cellcolor{oodgreen}60.11 &\cellcolor{oodgreen}42.27 &\cellcolor{oodgreen}78.31 &\cellcolor{oodgreen}\underline{66.10} &\cellcolor{oodgreen}75.89 & \cellcolor{oodgreen}38.74& \cellcolor{idred}90.33  & \cellcolor{idred}69.57 & \cellcolor{idred}84.04  & \cellcolor{idred}63.15 & \cellcolor{idred}81.94  & \cellcolor{idred}59.35 & \cellcolor{idred}70.40  & \cellcolor{idred}36.96 \\
        \midrule
        Ours (w/o Embodied) & \cellcolor{oodgreen}\textbf{94.40} &\cellcolor{oodgreen}\underline{67.98} &\cellcolor{oodgreen}\underline{77.85} &\cellcolor{oodgreen}\underline{53.47}&\cellcolor{oodgreen}71.31 &\cellcolor{oodgreen}\underline{58.05} &\cellcolor{oodgreen}60.67 &\cellcolor{oodgreen}\underline{52.34} &\cellcolor{oodgreen}78.68 &\cellcolor{oodgreen}63.79 &\cellcolor{oodgreen}\underline{77.57} & \cellcolor{oodgreen}\underline{52.56}& \cellcolor{oodgreen}89.96  & \cellcolor{oodgreen}\underline{83.68} &\cellcolor{oodgreen}\underline{85.00}&\cellcolor{oodgreen}\underline{72.63}&\cellcolor{oodgreen}79.96&\cellcolor{oodgreen}\underline{61.47}&\cellcolor{idred}70.08&\cellcolor{idred}\underline{46.38}\\
        Ours (w/o GUI) &\cellcolor{oodgreen}81.42 &\cellcolor{oodgreen}64.97 &\cellcolor{oodgreen}29.47 &\cellcolor{oodgreen}30.05 &\cellcolor{oodgreen}\underline{74.95} & \cellcolor{oodgreen}49.17 &\cellcolor{oodgreen}76.08 &\cellcolor{oodgreen}47.55 &\cellcolor{oodgreen}\underline{81.72} &\cellcolor{oodgreen}64.03 &\cellcolor{oodgreen}76.81 & \cellcolor{oodgreen}37.37& \cellcolor{oodgreen}\underline{91.56}  & \cellcolor{oodgreen}77.80 & \cellcolor{oodgreen}66.07  & \cellcolor{oodgreen}51.70 & \cellcolor{oodgreen}72.09  & \cellcolor{oodgreen}51.35 & \cellcolor{oodgreen}64.75  & \cellcolor{oodgreen}33.92 \\
        Ours &\cellcolor{oodgreen}\underline{93.90} &\cellcolor{oodgreen}\textbf{69.46} &\cellcolor{oodgreen}\textbf{78.15} &\cellcolor{oodgreen}\textbf{55.89} &\cellcolor{oodgreen}74.92 & \cellcolor{oodgreen}\textbf{59.72} &\cellcolor{oodgreen}\underline{64.39} &\cellcolor{oodgreen}\textbf{53.27} &\cellcolor{oodgreen}\textbf{82.54} &\cellcolor{oodgreen}\textbf{67.39} &\cellcolor{oodgreen}\textbf{81.14} & \cellcolor{oodgreen}\textbf{54.00}& \cellcolor{oodgreen}\textbf{91.95}  & \cellcolor{oodgreen}\textbf{86.17} & \cellcolor{oodgreen}\textbf{85.30}& \cellcolor{oodgreen}\textbf{72.99}& \cellcolor{oodgreen}\textbf{82.16} & \cellcolor{oodgreen}\textbf{62.47}& \cellcolor{idred}\textbf{73.60}& \cellcolor{idred}\textbf{48.35}\\
        \bottomrule
    \end{tabular}
     \vspace{-2mm}
    \caption{Results on different GUI tasks. The red background represents that the data source is in the training set of the corresponding model, while the green background represents that the test dataset is OOD for the model. \textbf{Bold} highlights the best results in the OOD setting, and underlined are the second-best.}
    \label{GUI performance}
    \vspace{-6mm}
\end{table*}

\subsection{Benchmarks and Metrics}

\textbf{GUI task.} For the evaluation of the GUI task, we employ seven distinct mobile, web, desktop and in-domain benchmarks: AC-High/Low~\cite{li2024effectsdatascaleui}, AITW~\cite{rawles2023androidinthewild}, GUIAct-P/W~\cite{chen2025guicoursegeneralvisionlanguage}, Llamatouch~\cite{zhang2024llamatouchfaithfulscalabletestbed}, AITZ~\cite{zhang2024androidzoochainofactionthoughtgui} OmniAct-W/D~\cite{kapoor2024omniactdatasetbenchmarkenabling} and Odyssey~\cite{lu2024guiodysseycomprehensivedataset}.
We follow OS-Atlas~\cite{wu2025osatlas} to take the success rate (SR) and grounding (GR) as the primary evaluation metrics. SR measures the per-step task success rate, while GR measures the accuracy of localizing the correct click coordinates.

We compare our model with the following methods: the proprietary GPT-4o~\cite{openai2024gpt4ocard}, SFT-based models such as OS-Atlas, Aguvis~\cite{xu2025aguvisunifiedpurevision} and Qwen2.5VL-7B* fine-tuned on our trajectory data, as well as RL-based models like GUI-R1~\cite{luo2025guir1generalistr1style}, infiGUI-R1~\cite{liu2025infiguir1advancingmultimodalgui}, UI-Shift~\cite{gao2025uishiftenhancingvlmbasedgui} and UI-AGILE~\cite{lian2025uiagileadvancingguiagents}. We note the concurrent work of OmniActor~\cite{yang2025omniactorgeneralistguiembodied}, which also proposes a unified agent for both 2D and 3D tasks. However, we exclude it from comparison because it is evaluated on a subset of benchmarks and its implementation is not publicly available.

\textbf{Embodied task.} 
We assess our model's performance through two distinct embodied tasks. First, we conduct spatial affordance prediction to assess the model's spatial grounding ability on the metric of SR. 
Specifically, we employ RoboReflT~\cite{lu2023vlgrasp6dofinteractivegrasp} for object referring and Where2Place~\cite{yuan2024robopointvisionlanguagemodelspatial}, RoboSpatial~\cite{song2025robospatialteachingspatialunderstanding}, RefSpatial~\cite{zhou2025roboreferspatialreferringreasoning} for free space referring. 
The second task is embodied navigation, which evaluates the model's practical application capabilities on the metric of SR and SPL (Success Rate Weighted by Inverse Path Length). We evaluate our approach on the unseen validation branch of ObjectNav~\cite{batra2020objectnavrevisitedevaluationembodied}. We adhere to the framework established in VLMNav~\cite{goetting2024endtoendnavigationvisionlanguage} and substitute the agent model in our experiments to assess the performance. We compare our model against the proprietary GPT-4o, open-source methods like Qwen2.5VL-7B, SpatialVLM like SpaceLLaVA~\cite{foutter2025spacellavavisionlanguagemodeladapted}, the latest method RoboPoint-13B~\cite{yuan2024robopointvisionlanguagemodelspatial}. More details of benchmarks and metrics are in Appendix~\ref{metrics details} and Appendix~\ref{embodied metrics details}.


\begin{figure*}[t]
    \centering
    \begin{subfigure}[b]{0.24\textwidth}
        \includegraphics[width=\linewidth]{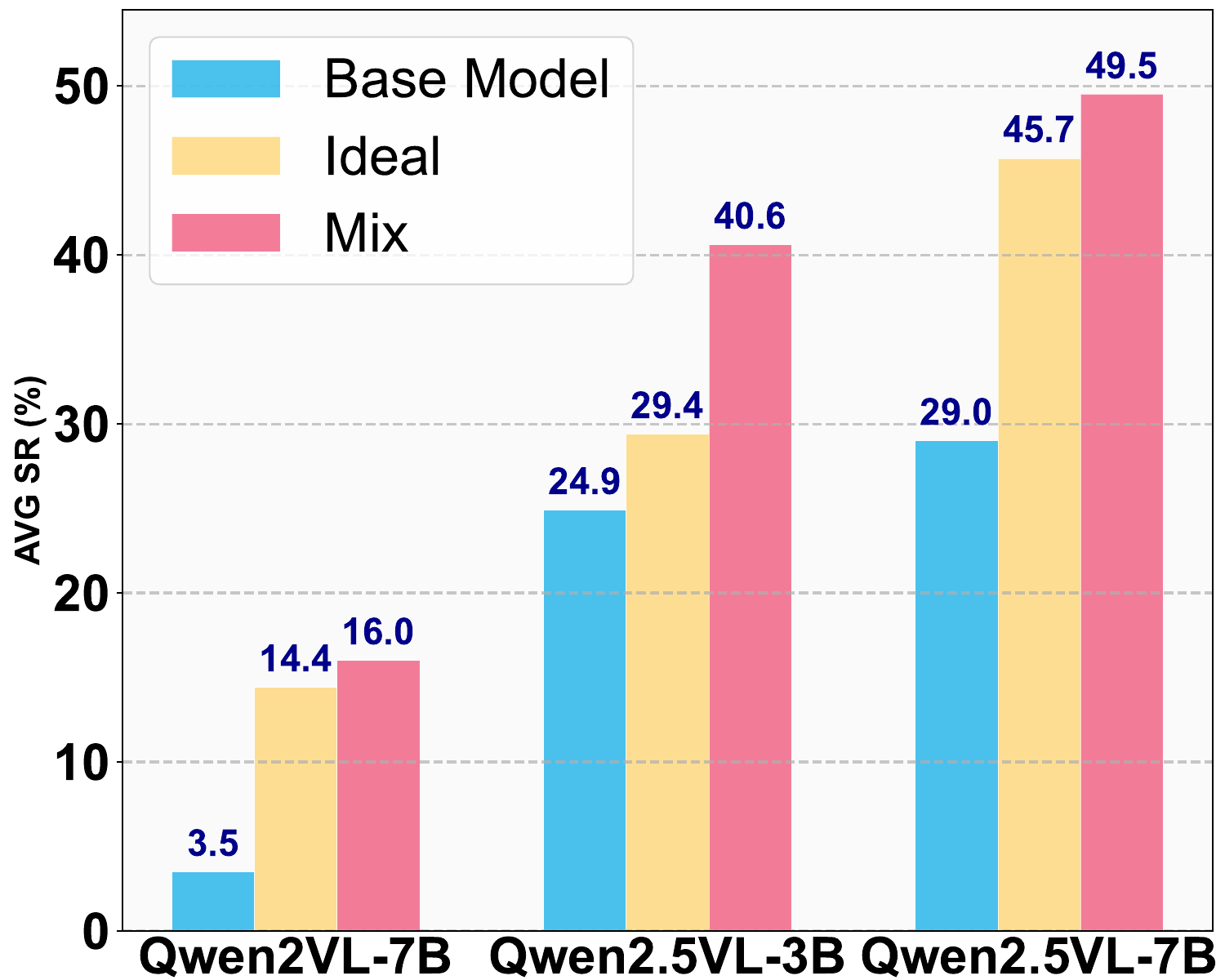}
        \vspace{-5mm}
        \caption{}
        \label{fig:basemodel}
    \end{subfigure}
    \hfill
    \begin{subfigure}[b]{0.24\textwidth}
        \includegraphics[width=\linewidth]{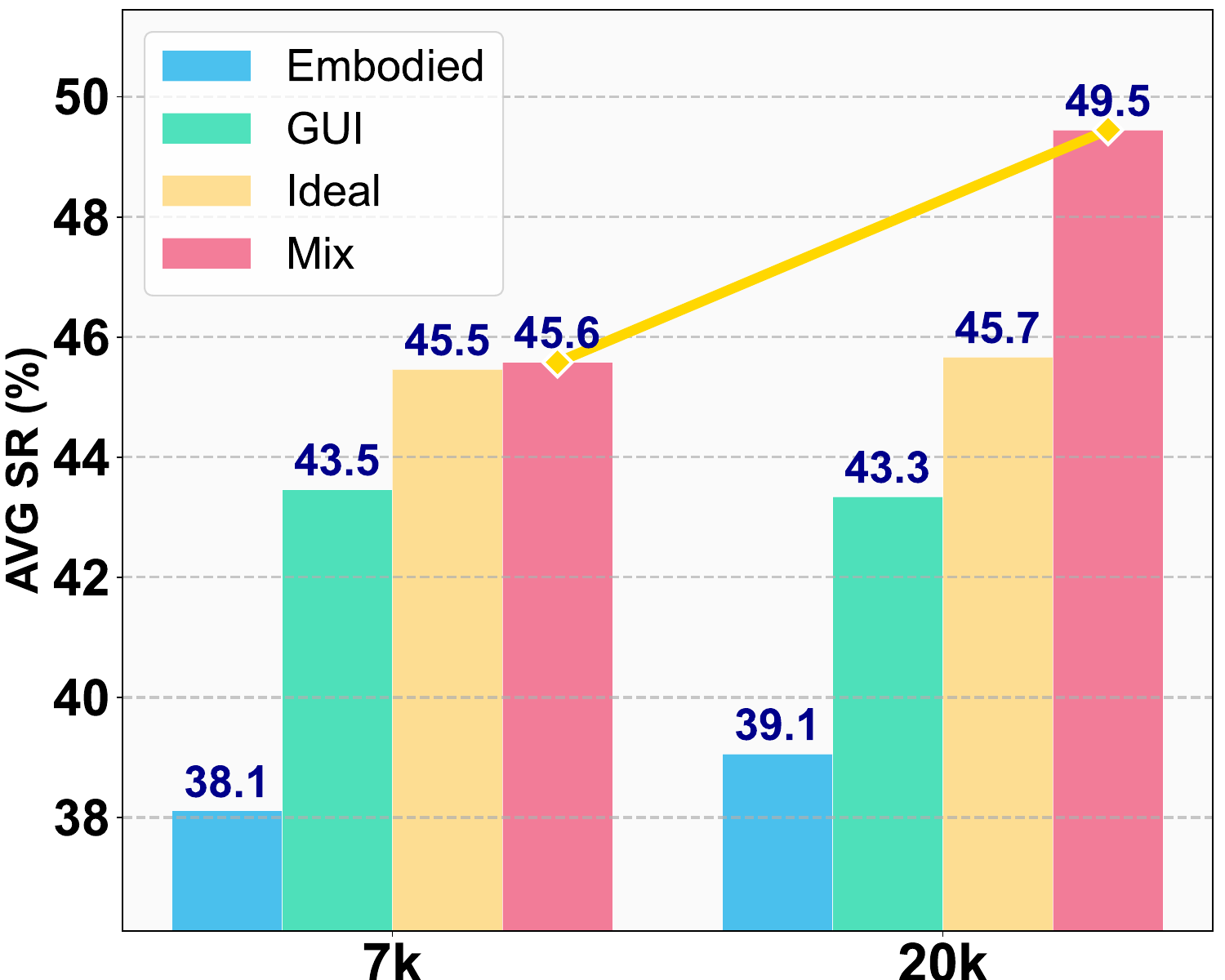}
        \vspace{-5mm}
        \caption{}
        \label{fig:data scale}
    \end{subfigure}
    \hfill
    \begin{subfigure}[b]{0.22\textwidth}
        \includegraphics[width=\linewidth]{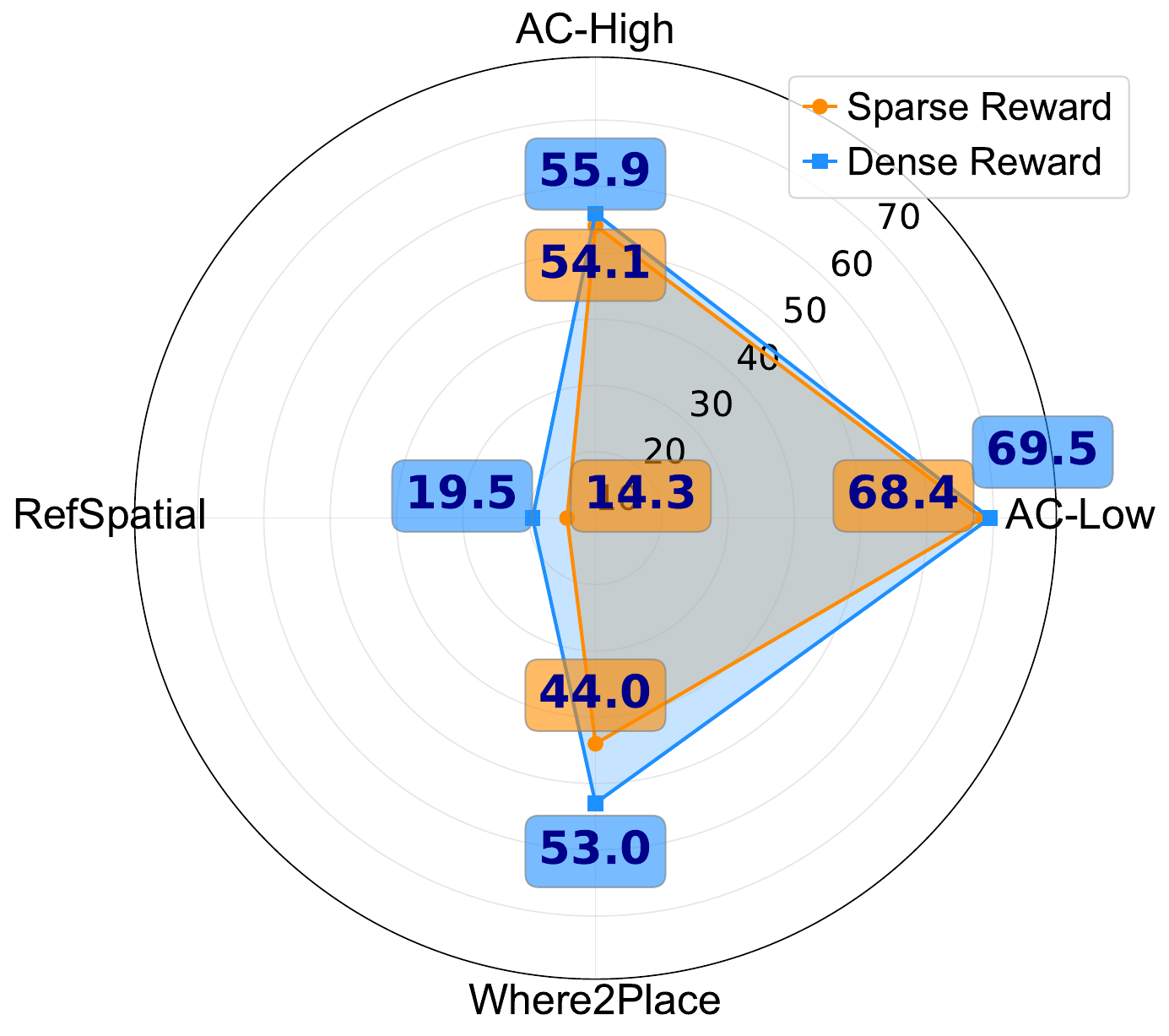} 
        \vspace{-5mm}
        \caption{}
        \label{fig:reward}
    \end{subfigure}
    \hfill
    \begin{subfigure}[b]{0.24\textwidth}
        \includegraphics[width=\linewidth]{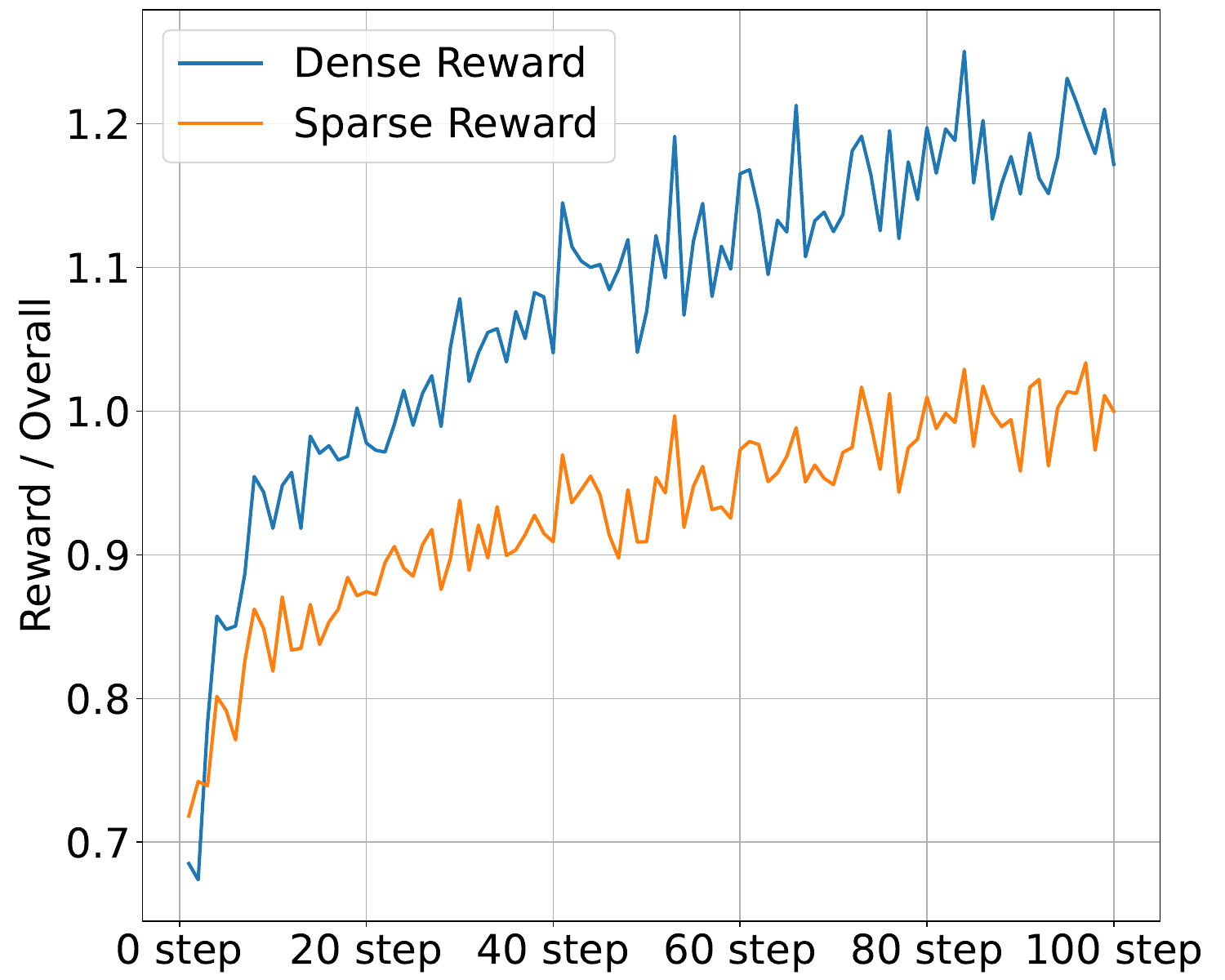} 
        \vspace{-5mm}
        \caption{}
        \label{fig:sube_row}
    \end{subfigure}
    \vspace{-3mm}
    \caption{Ablation Studies. (a) Performance of different base models. (b) Performance of different data scales. (c) Performance of dense and sparse reward. (d) Reward curves.}
    \label{fig:five_images_in_row}
    \vspace{-5mm}
\end{figure*}
\subsection{Main Results}
\textbf{GUI Navigation.} The results are shown in Table \ref{GUI performance}. To evaluate the generalization capability of NaviMaster, we employ entirely out-of-domain (OOD) test data, highlighted with a green background, which are disjoint from the training distribution. The remaining test cases, indicated with a red background, are in-domain. Compared with state-of-the-art baselines, NaviMaster demonstrates consistently superior performance across the various benchmarks, demonstrating strong generalization capability and robustness when handling OOD datasets.
Additionally, compared to models trained solely on either GUI or embodied data, our model trained on the mix data achieves the highest performance across all test datasets. This demonstrates the effectiveness of our collected visual-target trajectory and unified training framework, which enable strong generalization and competitive performance with a relatively small amount of data. 

\textbf{Spatial Affordance Prediction.}  
We evaluate two types of spatial affordance predicting datasets: one is object referring, which involves identifying and localizing specific objects within a scene based on language descriptions; the other is free space referring, which targets understanding and navigating to spatial regions or locations that are not necessarily tied to specific objects but are described in language. Table~\ref{results: spatial reference} summarizes the average success rate of predicted points falling within the ground-truth mask on the four spatial affordance prediction benchmarks. Compared to all baselines, NaviMaster performance best in all spatial affordance prediction tasks.
These results demonstrate that NaviMaster’s fine-grained visual–spatial alignment significantly enhances performance in both object-level and free-space referring.

\begin{table}[htbp]
\centering	
\tiny
\renewcommand{\arraystretch}{0.85} 
\begin{tabular}{l@{\hspace{3pt}}c@{\hspace{3pt}}c@{\hspace{3pt}}c@{\hspace{3pt}}c}
\toprule
     \bf{Models} &\bf{RoboReflt} &\bf{Where2Place}&\bf{RoboSpatial}&\bf{RefSpatial}\\
    \midrule
    GPT-4o &15.28 &29.00 &5.70 &8.40 \\
    Qwen2.5VL-7B & 3.46 & 3.00 & 10.31  & 3.55 \\
    SpaceLLaVA &21.30 &11.84 &2.50 & 4.02 \\
    RoboPoint-13B &49.82&46.77 &19.70 &8.40 \\
    \midrule
    Ours (w/o Embodied) &67.86& 35.05 & 14.75 & 16.88 \\
    Ours (w/o GUI) &76.23& 43.02 & 19.83 & 18.19 \\
    Ours &\bf{77.34}& \bf{52.97} & \bf{21.65} & \bf{19.49} \\
    \hline
\end{tabular}
 \vspace{-1mm}
\caption{Results on spatial affordance prediction.} 
\label{results: spatial reference}
\vspace{-3.5mm}
\end{table}

\textbf{Embodied Navigation.}  
Since we are the first to train a model capable of generalizing in VLMNav, there are no prior navigation models trained under VLMNav for direct comparison. We only report our results in Table~\ref{results: embodied navigation}, which reports performance on the ObjectNav benchmark. NaviMaster achieves the highest SR of 33.10\% and SPL of 12.60\%, representing a substantial improvement over the base model. Training exclusively on embodied data or GUI data yields slightly lower SR, indicating that the mix training strategy effectively leverages the complementary advantages of both data sources.
\begin{table}[htbp]
\centering	
\renewcommand{\arraystretch}{0.85} 
\tiny
\begin{tabular}{lcccccccccc}
\toprule
    \bf{Runs}&\bf{SR} &\bf{SPL}\\
    \midrule
    Qwen2.5VL-7B&27.23 &9.68\\
    RoboPoint   & 13.01    & 2.80   \\
    SpaceLLaVA   & 14.98    & 2.62  \\
    \midrule
    Ours (w/o GUI)&31.10&11.20\\
    Ours (w/o Embodied)&31.00&10.05\\
    Ours&\bf{33.20}&\bf{12.60}\\

    \hline
\end{tabular}
  \vspace{-1mm}
\caption{Results on embodied navigation.} 
\label{results: embodied navigation}
\end{table}

\subsection{Discussions and Analysis}
For all analysis experiments, we task AC-High/Low as GUI benchmarks and Where2Place, RefSpatial as Embodied benchmarks. More details and other ablation studies can be found in Appendix~\ref{Detailed Analysis}.

\begin{figure}[t]
    \centering
    \begin{subfigure}[b]{0.48\columnwidth}
        \includegraphics[width=\linewidth]{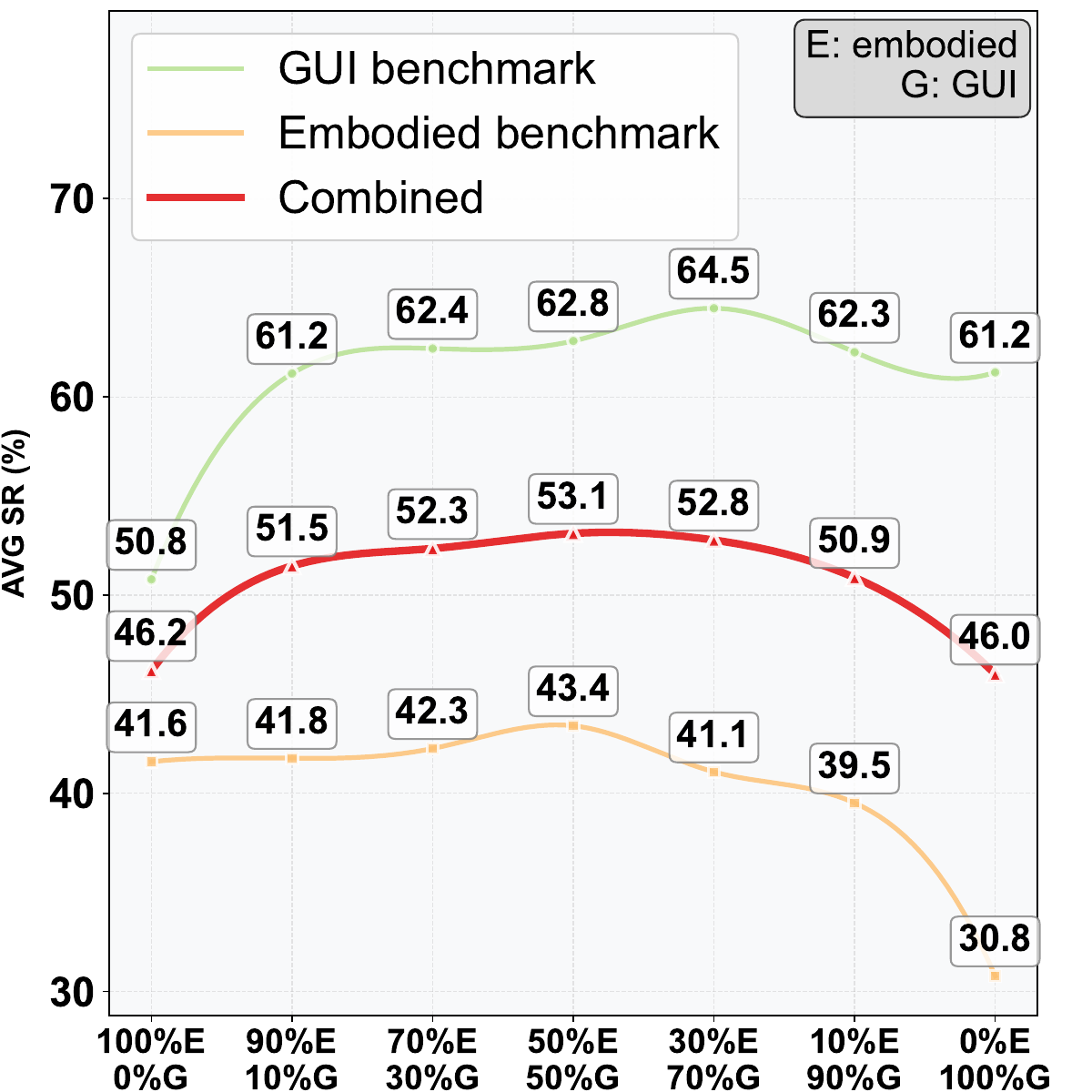}
        \vspace{-5mm}
        \caption{}
        \label{fig:ratio}
    \end{subfigure}
    \hfill
    \begin{subfigure}[b]{0.48\columnwidth}
        \includegraphics[width=\linewidth]{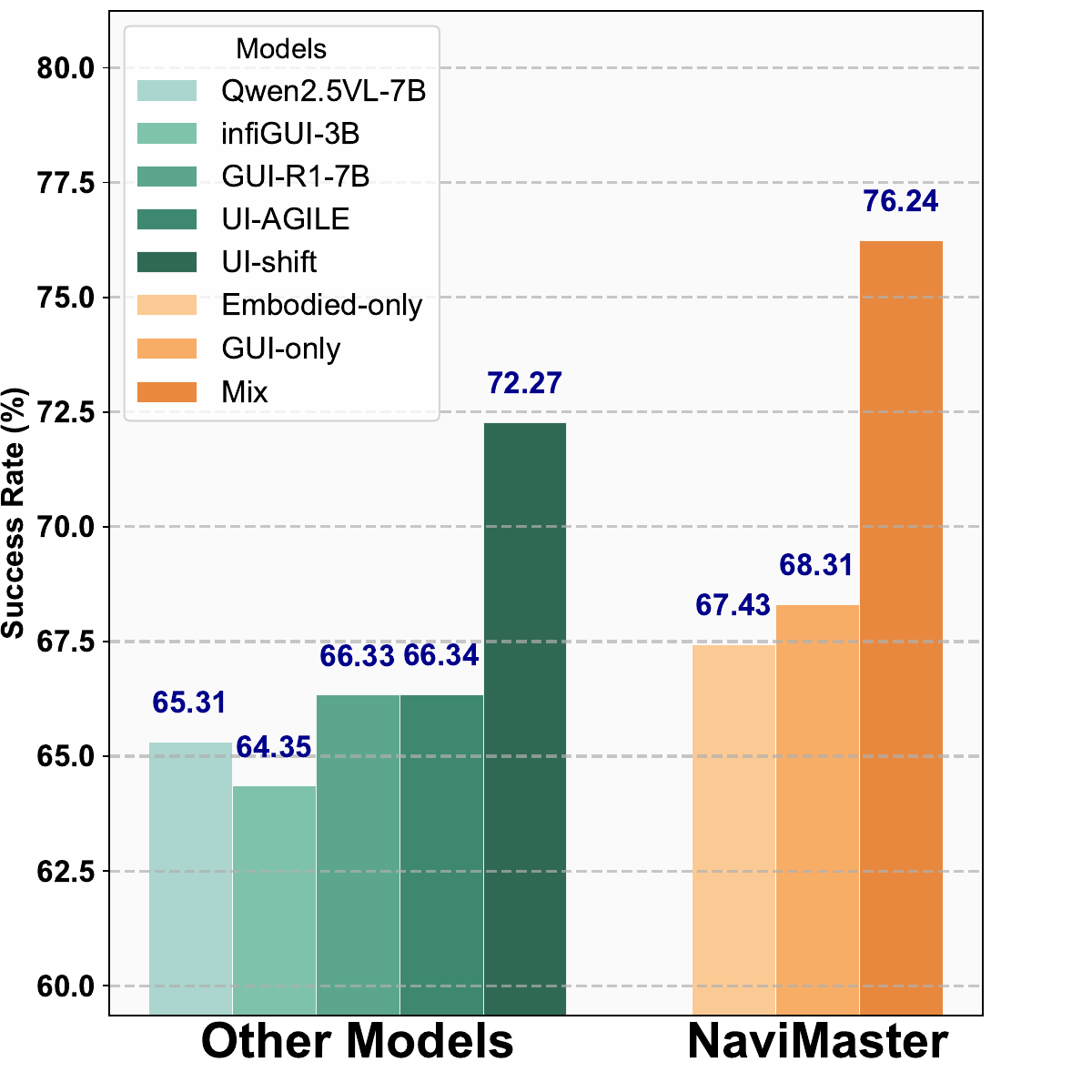}
        \vspace{-5mm}
        \caption{}
        \label{fig:moti}
    \end{subfigure}

    \vspace{-2mm}
    \caption{(a) Performance of different ratios of training data. (b) Performance on our visual-relation benchmark.}
    \label{fig:analysis}
    \vspace{-4mm}
\end{figure}

\textbf{Impact of data ratio.} We investigate the impact of varying data mixing ratios on performance across the two tasks. As illustrated in Fig.~\ref{fig:ratio}, we evaluate four distinct ratios (0:10, 1:9, 3:7 and 5:5) and report the average SR on both tasks. The results show that overall performance peaks at 5:5 ratio, demonstrating that our mixed training approach enhances cross-domain generalization. Notably, even with imbalanced mixtures, models trained on mix data outperform those trained on a single dataset.

\textbf{Advantages of mixing.} As stated in the introduction, to evaluate the visual reasoning enabled by our unified-training strategy, we manually constructed a visual-relation benchmark consisting of 100 instances from the AndroidControl dataset. Each instance contains a landmark UI element and a final target defined by its spatial relationship to that landmark. Importantly, the instructions require the model to infer the target location from a relational description (e.g., “Click on the 2nd button above the PUSH-UPS”) rather than directly naming the target. This design compels the model to reason over the image instead of merely learning a straightforward language-to-pixel mapping. As shown in Fig.~\ref{fig:moti}, the model trained with mix data achieves substantially higher accuracy than models trained on a single data type and also outperforms the existing RFT model with reasoning capabilities.


\textbf{Base model.} To verify that our improvements arise from unified training rather than a strong base model, we evaluated the framework on Qwen2.5VL-3B (smaller parameter scale) and Qwen2VL-7B~\cite{wang2024qwen2vlenhancingvisionlanguagemodels} (less pre-training knowledge). Fig.~\ref{fig:basemodel} presents the average success rate for each mode. Here, `Ideal' denotes the average success rates of two specialist models, each trained and evaluated solely within its respective domain (GUI or embodied). Training on mix data consistently outperforms training on single data, regardless of the underlying model.

\textbf{Data scales}. We investigate the impact of data scale on our mixture strategy's performance relative to single-task training. We evaluate two scales, 7k and 20k samples, while keeping the same training setting for all models. As shown in Fig.~\ref{fig:data scale}, the mix-data approach consistently outperforms single-task training, demonstrating its benefits in both large-data regimes and relatively few samples.

\textbf{Reward Design.} We assess the effectiveness of the proposed grounding dense reward by replacing it with a sparse alternative. Specifically, the threshold is set to $\hat{\theta_d} = 20$ in the sparse reward setting, whereas $\theta_d = 200$ is used in the dense reward setting. For the sparse reward, if the predicted point's distance to the ground-truth point is less than $\hat{\theta_d}$, the reward is 1; otherwise, it is 0. The results in the left of Fig.~\ref{fig:reward} show that the model trained with the dense reward consistently outperformed the one trained with the sparse reward. Moreover, the reward curve under dense setting in Fig.~\ref{fig:sube_row} rises more rapidly, indicating more efficient training.



\section{Conclusion}
We introduce NaviMaster, the first agent unifying GUI and embodied navigation in a single RL framework by reformulating both tasks into a visual-target trajectory format. This unification enables joint training and cross-task generalization. We propose a distance-aware dense reward to improve learning efficiency and spatial grounding capability. Experiments demonstrate NaviMaster achieves superior OOD generalization over prior methods.
\section*{Limitations}
While our NaviMaster improves the performance of both GUI and embodied navigation tasks significantly, especially in OOD scenes, it still treats GUI and embodied navigation as two different tasks. Our trajectory dataset lacks any single trajectory that interleaves GUI and embodied navigation tasks due to the collection difficulty. Future work should be constructing a navigation agent that supports interacting with the GUI and embodied environments at the same time. 

\section*{Broader Impacts} \label{risk}
The GUI or embodied data may leak personal information such as phone number or face. The data collection pipeline we proposed will not introduce any significant privacy such as personal information. It will not contain any real information as all the data source are virtual or from open-source datasets.

The navigation agent will interact with the OS system or real-world environments. This will potentially affect the functioning of the system or take risky actions to damage the environment. However, all the settings in our experiments are in virtual environments or on a monitor. We do not view this as a concern.

\section*{Acknowledgement}
This work was supported in part by the National Natural Science Foundation of China (62302167,  U23A20343, W2521174, and 62476090); in part by the Science
and Technology Commission of Shanghai (25511103300, 25511104302, 23ZR1420400); in part by Young Elite Scientists Sponsorship Program by CAST (YESS20240780).
\bibliography{main}

\begin{thebibliography}{51}
\providecommand{\natexlab}[1]{#1}

\bibitem[{Bai et~al.(2025)Bai, Chen, Liu, Wang, Ge, Song, Dang, Wang, Wang,
  Tang, Zhong, Zhu, Yang, Li, Wan, Wang, Ding, Fu, Xu, Ye, Zhang, Xie, Cheng,
  Zhang, Yang, Xu, and Lin}]{bai2025qwen25vltechnicalreport}
Shuai Bai, Keqin Chen, Xuejing Liu, Jialin Wang, Wenbin Ge, Sibo Song, Kai
  Dang, Peng Wang, Shijie Wang, Jun Tang, Humen Zhong, Yuanzhi Zhu, Mingkun
  Yang, Zhaohai Li, Jianqiang Wan, Pengfei Wang, Wei Ding, Zheren Fu, Yiheng
  Xu, and 8 others. 2025.
\newblock \href {https://arxiv.org/abs/2502.13923} {Qwen2.5-vl technical
  report}.
\newblock \emph{Preprint}, arXiv:2502.13923.

\bibitem[{Batra et~al.(2020)Batra, Gokaslan, Kembhavi, Maksymets, Mottaghi,
  Savva, Toshev, and Wijmans}]{batra2020objectnavrevisitedevaluationembodied}
Dhruv Batra, Aaron Gokaslan, Aniruddha Kembhavi, Oleksandr Maksymets, Roozbeh
  Mottaghi, Manolis Savva, Alexander Toshev, and Erik Wijmans. 2020.
\newblock \href {https://arxiv.org/abs/2006.13171} {Objectnav revisited: On
  evaluation of embodied agents navigating to objects}.
\newblock \emph{Preprint}, arXiv:2006.13171.

\bibitem[{Chen et~al.(2025{\natexlab{a}})Chen, Ji, Zhong, Zhu, Li, Gan, Huang,
  Zou, Liu, Chen, Chen, and Shen}]{chen2025guishepherdreliableprocessreward}
Cong Chen, Kaixiang Ji, Hao Zhong, Muzhi Zhu, Anzhou Li, Guo Gan, Ziyuan Huang,
  Cheng Zou, Jiajia Liu, Jingdong Chen, Hao Chen, and Chunhua Shen.
  2025{\natexlab{a}}.
\newblock \href {https://arxiv.org/abs/2509.23738} {Gui-shepherd: Reliable
  process reward and verification for long-sequence gui tasks}.
\newblock \emph{Preprint}, arXiv:2509.23738.

\bibitem[{Chen et~al.(2025{\natexlab{b}})Chen, Cui, Hu, Qin, Fang, Zhao, Wang,
  Liu, Chen, Huo et~al.}]{chen2025guicoursegeneralvisionlanguage}
Wentong Chen, Junbo Cui, Jinyi Hu, Yujia Qin, Junjie Fang, Yue Zhao, Chongyi
  Wang, Jun Liu, Guirong Chen, Yupeng Huo, and 1 others. 2025{\natexlab{b}}.
\newblock Guicourse: From general vision language model to versatile gui agent.
\newblock In \emph{Proceedings of the 63rd Annual Meeting of the Association
  for Computational Linguistics (Volume 1: Long Papers)}, pages 21936--21959.

\bibitem[{Foutter et~al.(2025)Foutter, Gammelli, Kruger, Foss, Bhoj, Guffanti,
  D'Amico, and Pavone}]{foutter2025spacellavavisionlanguagemodeladapted}
Matthew Foutter, Daniele Gammelli, Justin Kruger, Ethan Foss, Praneet Bhoj,
  Tommaso Guffanti, Simone D'Amico, and Marco Pavone. 2025.
\newblock Space-llava: A vision-language model adapted to extraterrestrial
  applications.
\newblock In \emph{2025 IEEE Aerospace Conference}, pages 1--23. IEEE.

\bibitem[{Gan et~al.(2026)Gan, Ding, Chen, Ren, Huang, and
  Zhou}]{gan2026androidcoachimproveonline}
Guo Gan, Yuxuan Ding, Cong Chen, Yuwei Ren, Yin Huang, and Hong Zhou. 2026.
\newblock \href {https://arxiv.org/abs/2604.07277} {Android coach: Improve
  online agentic training efficiency with single state multiple actions}.
\newblock \emph{Preprint}, arXiv:2604.07277.

\bibitem[{Gao et~al.(2026{\natexlab{a}})Gao, Zhang, Gao, Liu, Luan, and
  Xu}]{gao2025uishiftenhancingvlmbasedgui}
Longxi Gao, Li~Zhang, Pengzhi Gao, Wei Liu, Jian Luan, and Mengwei Xu.
  2026{\natexlab{a}}.
\newblock \href {https://openreview.net/forum?id=NakMHPljT7} {{GUI}-shift:
  Enhancing {VLM}-based {GUI} agents through self-supervised reinforcement
  learning}.
\newblock In \emph{The Fourteenth International Conference on Learning
  Representations}.

\bibitem[{Gao et~al.(2024)Gao, Wang, Gao, Wang, and
  Yuan}]{gao2024visionlanguagenavigationembodiedintelligence}
Peng Gao, Peng Wang, Feng Gao, Fei Wang, and Ruyue Yuan. 2024.
\newblock \href {https://arxiv.org/abs/2402.14304} {Vision-language navigation
  with embodied intelligence: A survey}.
\newblock \emph{Preprint}, arXiv:2402.14304.

\bibitem[{Gao et~al.(2026{\natexlab{b}})Gao, Wang, Tan, and Xie}]{gao2026tpru}
Zhenkun Gao, Xuhong Wang, Xin Tan, and Yuan Xie. 2026{\natexlab{b}}.
\newblock \href {https://openreview.net/forum?id=crOvAD9MPA} {Tpru: Advancing
  temporal and procedural understanding in large multimodal models}.
\newblock In \emph{The Fourteenth International Conference on Learning
  Representations}.

\bibitem[{Goetting et~al.(2025)Goetting, Singh, and
  Loquercio}]{goetting2024endtoendnavigationvisionlanguage}
Dylan Goetting, Himanshu~Gaurav Singh, and Antonio Loquercio. 2025.
\newblock End-to-end navigation with vision-language models: Transforming
  spatial reasoning into question-answering.
\newblock In \emph{International Conference on Neuro-symbolic Systems}, pages
  22--35. PMLR.

\bibitem[{Hart et~al.(1968)Hart, Nilsson, and Raphael}]{4082128}
Peter~E. Hart, Nils~J. Nilsson, and Bertram Raphael. 1968.
\newblock \href {https://doi.org/10.1109/TSSC.1968.300136} {A formal basis for
  the heuristic determination of minimum cost paths}.
\newblock \emph{IEEE Transactions on Systems Science and Cybernetics},
  4(2):100--107.

\bibitem[{Hong et~al.(2025)Hong, Sun, Li, Yao, Wu, Chien, Yin, Wu, Wang, and
  Chang}]{hong2025embodiedwebagentsbridging}
Yining Hong, Rui Sun, Bingxuan Li, Xingcheng Yao, Maxine Wu, Alexander Chien,
  Da~Yin, Ying~Nian Wu, Zhecan Wang, and Kai-Wei Chang. 2025.
\newblock Embodied web agents: Bridging physical-digital realms for integrated
  agent intelligence.
\newblock In \emph{The Thirty-ninth Annual Conference on Neural Information
  Processing Systems Datasets and Benchmarks Track}.

\bibitem[{Ji et~al.(2026)Ji, Tian, Zhu, Jiang, Cao, Ma, Xie, and
  Tan}]{ji2026s2dsparsedenselifting}
Yuzhou Ji, Qijian Tian, He~Zhu, Xiaoqi Jiang, Guangzhi Cao, Lizhuang Ma, Yuan
  Xie, and Xin Tan. 2026.
\newblock \href {https://arxiv.org/abs/2603.10893} {S2d: Sparse to dense
  lifting for 3d reconstruction with minimal inputs}.

\bibitem[{Ji et~al.(2025)Ji, Zhu, Tang, Liu, Zhang, Tan, and
  Xie}]{ji2025fastlgs}
Yuzhou Ji, He~Zhu, Junshu Tang, Wuyi Liu, Zhizhong Zhang, Xin Tan, and Yuan
  Xie. 2025.
\newblock Fastlgs: Speeding up language embedded gaussians with feature grid
  mapping.
\newblock In \emph{Proceedings of the AAAI conference on artificial
  intelligence}, volume~39, pages 3922--3930.

\bibitem[{Jin et~al.(2025)Jin, Li, Gu, Liu, Zhao, Lai, Gan, Wang, Wang, Tan,
  and Ma}]{Jin_2025}
Yizhang Jin, Jian Li, Tianjun Gu, Yexin Liu, Bo~Zhao, Jinxiang Lai, Zhenye Gan,
  Yabiao Wang, Chengjie Wang, Xin Tan, and Lizhuang Ma. 2025.
\newblock \href {https://doi.org/10.1007/s44267-025-00099-6} {Efficient
  multimodal large language models: a survey}.
\newblock \emph{Visual Intelligence}, 3(1).

\bibitem[{Kapoor et~al.(2024)Kapoor, Butala, Russak, Koh, Kamble, AlShikh, and
  Salakhutdinov}]{kapoor2024omniactdatasetbenchmarkenabling}
Raghav Kapoor, Yash~Parag Butala, Melisa Russak, Jing~Yu Koh, Kiran Kamble,
  Waseem AlShikh, and Ruslan Salakhutdinov. 2024.
\newblock Omniact: A dataset and benchmark for enabling multimodal generalist
  autonomous agents for desktop and web.
\newblock In \emph{European Conference on Computer Vision}, pages 161--178.
  Springer.

\bibitem[{Kwon et~al.(2023)Kwon, Li, Zhuang, Sheng, Zheng, Yu, Gonzalez, Zhang,
  and Stoica}]{kwon2023efficient}
Woosuk Kwon, Zhuohan Li, Siyuan Zhuang, Ying Sheng, Lianmin Zheng, Cody~Hao Yu,
  Joseph~E. Gonzalez, Hao Zhang, and Ion Stoica. 2023.
\newblock Efficient memory management for large language model serving with
  pagedattention.
\newblock In \emph{Proceedings of the ACM SIGOPS 29th Symposium on Operating
  Systems Principles}.

\bibitem[{Li et~al.(2026)Li, Zhao, Xie, Tan, and Li}]{li2026compassnav}
LinFeng Li, Jian Zhao, Yuan Xie, Xin Tan, and Xuelong Li. 2026.
\newblock \href {https://openreview.net/forum?id=eqcDckWHik} {Compassnav:
  Steering from path imitation to decision understanding in navigation}.
\newblock In \emph{The Fourteenth International Conference on Learning
  Representations}.

\bibitem[{Li et~al.(2024)Li, Bishop, Li, Rawles, Campbell-Ajala, Tyamagundlu,
  and Riva}]{li2024effectsdatascaleui}
Wei Li, William Bishop, Alice Li, Chris Rawles, Folawiyo Campbell-Ajala, Divya
  Tyamagundlu, and Oriana Riva. 2024.
\newblock On the effects of data scale on ui control agents.
\newblock \emph{Advances in Neural Information Processing Systems},
  37:92130--92154.

\bibitem[{Lian et~al.(2025)Lian, Wu, Ma, Ding, Song, Chen, Zheng, and
  Li}]{lian2025uiagileadvancingguiagents}
Shuquan Lian, Yuhang Wu, Jia Ma, Yifan Ding, Zihan Song, Bingqi Chen, Xiawu
  Zheng, and Hui Li. 2025.
\newblock \href {https://arxiv.org/abs/2507.22025} {Ui-agile: Advancing gui
  agents with effective reinforcement learning and precise inference-time
  grounding}.
\newblock \emph{Preprint}, arXiv:2507.22025.

\bibitem[{Lin et~al.(2025)Lin, Nie, Zai, Wei, Han, Xu, Niu, Han, Lin, Lu, and
  Liang}]{lin2025evolvenavselfimprovingembodiedreasoning}
Bingqian Lin, Yunshuang Nie, Khun~Loun Zai, Ziming Wei, Mingfei Han, Rongtao
  Xu, Minzhe Niu, Jianhua Han, Liang Lin, Cewu Lu, and Xiaodan Liang. 2025.
\newblock \href {https://arxiv.org/abs/2506.01551} {Evolvenav: Self-improving
  embodied reasoning for llm-based vision-language navigation}.
\newblock \emph{Preprint}, arXiv:2506.01551.

\bibitem[{Liu et~al.(2025{\natexlab{a}})Liu, Bao, Lin, Wang, Tan, Wang, Xie,
  and Lu}]{liu2025idmr}
Bangwei Liu, Yicheng Bao, Shaohui Lin, Xuhong Wang, Xin Tan, Yingchun Wang,
  Yuan Xie, and Chaochao Lu. 2025{\natexlab{a}}.
\newblock Idmr: Towards instance-driven precise visual correspondence in
  multimodal retrieval.
\newblock In \emph{Proceedings of the IEEE/CVF International Conference on
  Computer Vision}, pages 6320--6329.

\bibitem[{Liu et~al.(2025{\natexlab{b}})Liu, Li, Xie, Hu, Han, Zhang, Yang, and
  Wu}]{liu2025infiguir1advancingmultimodalgui}
Yuhang Liu, Pengxiang Li, Congkai Xie, Xavier Hu, Xiaotian Han, Shengyu Zhang,
  Hongxia Yang, and Fei Wu. 2025{\natexlab{b}}.
\newblock \href {https://arxiv.org/abs/2504.14239} {Infigui-r1: Advancing
  multimodal gui agents from reactive actors to deliberative reasoners}.
\newblock \emph{Preprint}, arXiv:2504.14239.

\bibitem[{Liu et~al.(2025{\natexlab{c}})Liu, Sun, Zang, Dong, Cao, Duan, Lin,
  and Wang}]{liu2025visualrftvisualreinforcementfinetuning}
Ziyu Liu, Zeyi Sun, Yuhang Zang, Xiaoyi Dong, Yuhang Cao, Haodong Duan, Dahua
  Lin, and Jiaqi Wang. 2025{\natexlab{c}}.
\newblock Visual-rft: Visual reinforcement fine-tuning.
\newblock In \emph{Proceedings of the IEEE/CVF International Conference on
  Computer Vision}, pages 2034--2044.

\bibitem[{Lu et~al.(2025{\natexlab{a}})Lu, Shao, Liu, Du, Meng, Li, Chen,
  Huang, Zhang, and Luo}]{lu2024guiodysseycomprehensivedataset}
Quanfeng Lu, Wenqi Shao, Zitao Liu, Lingxiao Du, Fanqing Meng, Boxuan Li,
  Botong Chen, Siyuan Huang, Kaipeng Zhang, and Ping Luo. 2025{\natexlab{a}}.
\newblock Guiodyssey: A comprehensive dataset for cross-app gui navigation on
  mobile devices.
\newblock In \emph{Proceedings of the IEEE/CVF International Conference on
  Computer Vision}, pages 22404--22414.

\bibitem[{Lu et~al.(2023)Lu, Fan, Deng, Liu, Li, and
  Wang}]{lu2023vlgrasp6dofinteractivegrasp}
Yuhao Lu, Yixuan Fan, Beixing Deng, Fangfu Liu, Yali Li, and Shengjin Wang.
  2023.
\newblock Vl-grasp: a 6-dof interactive grasp policy for language-oriented
  objects in cluttered indoor scenes.
\newblock In \emph{2023 IEEE/RSJ International Conference on Intelligent Robots
  and Systems (IROS)}, pages 976--983. IEEE.

\bibitem[{Lu et~al.(2025{\natexlab{b}})Lu, Chai, Guo, Yin, Liu, Wang, Xiao,
  Ren, Xiong, and Li}]{lu2025uir1enhancingefficientaction}
Zhengxi Lu, Yuxiang Chai, Yaxuan Guo, Xi~Yin, Liang Liu, Hao Wang, Han Xiao,
  Shuai Ren, Guanjing Xiong, and Hongsheng Li. 2025{\natexlab{b}}.
\newblock \href {https://arxiv.org/abs/2503.21620} {Ui-r1: Enhancing efficient
  action prediction of gui agents by reinforcement learning}.
\newblock \emph{Preprint}, arXiv:2503.21620.

\bibitem[{Luo et~al.(2025)Luo, Wang, He, and
  Xia}]{luo2025guir1generalistr1style}
Run Luo, Lu~Wang, Wanwei He, and Xiaobo Xia. 2025.
\newblock \href {https://arxiv.org/abs/2504.10458} {Gui-r1 : A generalist
  r1-style vision-language action model for gui agents}.
\newblock \emph{Preprint}, arXiv:2504.10458.

\bibitem[{OpenAI et~al.(2024)OpenAI, :, Hurst, Lerer, Goucher, Perelman,
  Ramesh, Clark, Ostrow, Welihinda, Hayes, Radford, Mądry, Baker-Whitcomb,
  Beutel, Borzunov, Carney, Chow, Kirillov, Nichol, Paino, Renzin, Passos,
  Kirillov, Christakis, Conneau, Kamali, Jabri, Moyer, Tam, Crookes,
  Tootoochian, Tootoonchian, Kumar, Vallone, Karpathy, Braunstein, Cann,
  Codispoti, Galu, Kondrich, Tulloch, Mishchenko, Baek, Jiang, Pelisse,
  Woodford, Gosalia, Dhar, Pantuliano, Nayak, Oliver, Zoph, Ghorbani,
  Leimberger, Rossen, Sokolowsky, Wang, Zweig, Hoover, Samic, McGrew, Spero,
  Giertler, Cheng, Lightcap, Walkin, Quinn, Guarraci, Hsu, Kellogg, Eastman,
  Lugaresi, Wainwright, Bassin, Hudson, Chu, Nelson, Li, Shern, Conger,
  Barette, Voss, Ding, Lu, Zhang, Beaumont, Hallacy, Koch, Gibson, Kim, Choi,
  McLeavey, Hesse, Fischer, Winter, Czarnecki, Jarvis, Wei, Koumouzelis,
  Sherburn, Kappler, Levin, Levy, Carr, Farhi, Mely, Robinson, Sasaki, Jin,
  Valladares, Tsipras, Li, Nguyen, Findlay, Oiwoh, Wong, Asdar, Proehl, Yang,
  Antonow, Kramer, Peterson, Sigler, Wallace, Brevdo, Mays, Khorasani, Such,
  Raso, Zhang, von Lohmann, Sulit, Goh, Oden, Salmon, Starace, Brockman,
  Salman, Bao, Hu, Wong, Wang, Schmidt, Whitney, Jun, Kirchner,
  de~Oliveira~Pinto, Ren, Chang, Chung, Kivlichan, O'Connell, O'Connell,
  Osband, Silber, Sohl, Okuyucu, Lan, Kostrikov, Sutskever, Kanitscheider,
  Gulrajani, Coxon, Menick, Pachocki, Aung, Betker, Crooks, Lennon, Kiros,
  Leike, Park, Kwon, Phang, Teplitz, Wei, Wolfe, Chen, Harris, Varavva, Lee,
  Shieh, Lin, Yu, Weng, Tang, Yu, Jang, Candela, Beutler, Landers, Parish,
  Heidecke, Schulman, Lachman, McKay, Uesato, Ward, Kim, Huizinga, Sitkin,
  Kraaijeveld, Gross, Kaplan, Snyder, Achiam, Jiao, Lee, Zhuang, Harriman,
  Fricke, Hayashi, Singhal, Shi, Karthik, Wood, Rimbach, Hsu, Nguyen,
  Gu-Lemberg, Button, Liu, Howe, Muthukumar, Luther, Ahmad, Kai, Itow, Workman,
  Pathak, Chen, Jing, Guy, Fedus, Zhou, Mamitsuka, Weng, McCallum, Held,
  Ouyang, Feuvrier, Zhang, Kondraciuk, Kaiser, Hewitt, Metz, Doshi, Aflak,
  Simens, Boyd, Thompson, Dukhan, Chen, Gray, Hudnall, Zhang, Aljubeh, Litwin,
  Zeng, Johnson, Shetty, Gupta, Shah, Yatbaz, Yang, Zhong, Glaese, Chen,
  Janner, Lampe, Petrov, Wu, Wang, Fradin, Pokrass, Castro, de~Castro, Pavlov,
  Brundage, Wang, Khan, Murati, Bavarian, Lin, Yesildal, Soto, Gimelshein,
  Cone, Staudacher, Summers, LaFontaine, Chowdhury, Ryder, Stathas, Turley,
  Tezak, Felix, Kudige, Keskar, Deutsch, Bundick, Puckett, Nachum, Okelola,
  Boiko, Murk, Jaffe, Watkins, Godement, Campbell-Moore, Chao, McMillan, Belov,
  Su, Bak, Bakkum, Deng, Dolan, Hoeschele, Welinder, Tillet, Pronin, Tillet,
  Dhariwal, Yuan, Dias, Lim, Arora, Troll, Lin, Lopes, Puri, Miyara, Leike,
  Gaubert, Zamani, Wang, Donnelly, Honsby, Smith, Sahai, Ramchandani, Huet,
  Carmichael, Zellers, Chen, Chen, Nigmatullin, Cheu, Jain, Altman, Schoenholz,
  Toizer, Miserendino, Agarwal, Culver, Ethersmith, Gray, Grove, Metzger,
  Hermani, Jain, Zhao, Wu, Jomoto, Wu, Shuaiqi, Xia, Phene, Papay, Narayanan,
  Coffey, Lee, Hall, Balaji, Broda, Stramer, Xu, Gogineni, Christianson,
  Sanders, Patwardhan, Cunninghman, Degry, Dimson, Raoux, Shadwell, Zheng,
  Underwood, Markov, Sherbakov, Rubin, Stasi, Kaftan, Heywood, Peterson,
  Walters, Eloundou, Qi, Moeller, Monaco, Kuo, Fomenko, Chang, Zheng, Zhou,
  Manassra, Sheu, Zaremba, Patil, Qian, Kim, Cheng, Zhang, He, Zhang, Jin, Dai,
  and Malkov}]{openai2024gpt4ocard}
OpenAI, :, Aaron Hurst, Adam Lerer, Adam~P. Goucher, Adam Perelman, Aditya
  Ramesh, Aidan Clark, AJ~Ostrow, Akila Welihinda, Alan Hayes, Alec Radford,
  Aleksander Mądry, Alex Baker-Whitcomb, Alex Beutel, Alex Borzunov, Alex
  Carney, Alex Chow, Alex Kirillov, and 401 others. 2024.
\newblock \href {https://arxiv.org/abs/2410.21276} {Gpt-4o system card}.
\newblock \emph{Preprint}, arXiv:2410.21276.

\bibitem[{Qin et~al.(2025)Qin, Ye, Fang, Wang, Liang, Tian, Zhang, Li, Li,
  Huang, Zhong, Li, Yang, Miao, Lin, Liu, Jiang, Ma, Li, Xiao, Cai, Li, Zheng,
  Jin, Li, Zhou, Wang, Chen, Li, Yang, Liu, Lin, Peng, Liu, and
  Shi}]{qin2025uitarspioneeringautomatedgui}
Yujia Qin, Yining Ye, Junjie Fang, Haoming Wang, Shihao Liang, Shizuo Tian,
  Junda Zhang, Jiahao Li, Yunxin Li, Shijue Huang, Wanjun Zhong, Kuanye Li,
  Jiale Yang, Yu~Miao, Woyu Lin, Longxiang Liu, Xu~Jiang, Qianli Ma, Jingyu Li,
  and 16 others. 2025.
\newblock \href {https://arxiv.org/abs/2501.12326} {Ui-tars: Pioneering
  automated gui interaction with native agents}.
\newblock \emph{Preprint}, arXiv:2501.12326.

\bibitem[{Ramakrishnan et~al.(2021)Ramakrishnan, Gokaslan, Wijmans, Maksymets,
  Clegg, Turner, Undersander, Galuba, Westbury, Chang
  et~al.}]{ramakrishnan2021hm3d}
Santhosh~Kumar Ramakrishnan, Aaron Gokaslan, Erik Wijmans, Oleksandr Maksymets,
  Alexander Clegg, John~M Turner, Eric Undersander, Wojciech Galuba, Andrew
  Westbury, Angel~X Chang, and 1 others. 2021.
\newblock Habitat-matterport 3d dataset (hm3d): 1000 large-scale 3d
  environments for embodied ai.
\newblock In \emph{Thirty-fifth Conference on Neural Information Processing
  Systems Datasets and Benchmarks Track (Round 2)}.

\bibitem[{Rawles et~al.(2023)Rawles, Li, Rodriguez, Riva, and
  Lillicrap}]{rawles2023androidinthewild}
Christopher Rawles, Alice Li, Daniel Rodriguez, Oriana Riva, and Timothy~P
  Lillicrap. 2023.
\newblock \href {https://openreview.net/forum?id=j4b3l5kOil} {Androidinthewild:
  A large-scale dataset for android device control}.
\newblock In \emph{Thirty-seventh Conference on Neural Information Processing
  Systems Datasets and Benchmarks Track}.

\bibitem[{Savva et~al.(2019)Savva, Kadian, Maksymets, Zhao, Wijmans, Jain,
  Straub, Liu, Koltun, Malik et~al.}]{savva2019habitatplatformembodiedai}
Manolis Savva, Abhishek Kadian, Oleksandr Maksymets, Yili Zhao, Erik Wijmans,
  Bhavana Jain, Julian Straub, Jia Liu, Vladlen Koltun, Jitendra Malik, and 1
  others. 2019.
\newblock Habitat: A platform for embodied ai research.
\newblock In \emph{Proceedings of the IEEE/CVF international conference on
  computer vision}, pages 9339--9347.

\bibitem[{Song et~al.(2025)Song, Blukis, Tremblay, Tyree, Su, and
  Birchfield}]{song2025robospatialteachingspatialunderstanding}
Chan~Hee Song, Valts Blukis, Jonathan Tremblay, Stephen Tyree, Yu~Su, and Stan
  Birchfield. 2025.
\newblock Robospatial: Teaching spatial understanding to 2d and 3d
  vision-language models for robotics.
\newblock In \emph{Proceedings of the Computer Vision and Pattern Recognition
  Conference}, pages 15768--15780.

\bibitem[{Tian et~al.(2025)Tian, Tan, Xie, and Ma}]{tian2025drivingforward}
Qijian Tian, Xin Tan, Yuan Xie, and Lizhuang Ma. 2025.
\newblock Drivingforward: Feed-forward 3d gaussian splatting for driving scene
  reconstruction from flexible surround-view input.
\newblock In \emph{Proceedings of the AAAI Conference on Artificial
  Intelligence}, volume~39, pages 7374--7382.

\bibitem[{Wang et~al.(2024)Wang, Bai, Tan, Wang, Fan, Bai, Chen, Liu, Wang, Ge,
  Fan, Dang, Du, Ren, Men, Liu, Zhou, Zhou, and
  Lin}]{wang2024qwen2vlenhancingvisionlanguagemodels}
Peng Wang, Shuai Bai, Sinan Tan, Shijie Wang, Zhihao Fan, Jinze Bai, Keqin
  Chen, Xuejing Liu, Jialin Wang, Wenbin Ge, Yang Fan, Kai Dang, Mengfei Du,
  Xuancheng Ren, Rui Men, Dayiheng Liu, Chang Zhou, Jingren Zhou, and Junyang
  Lin. 2024.
\newblock \href {https://arxiv.org/abs/2409.12191} {Qwen2-vl: Enhancing
  vision-language model's perception of the world at any resolution}.
\newblock \emph{Preprint}, arXiv:2409.12191.

\bibitem[{Wang et~al.(2026)Wang, Liu, Gao, Ma, Wang, Xie, and
  Tan}]{wang2026explorelongtermmemorybenchmark}
Sen Wang, Bangwei Liu, Zhenkun Gao, Lizhuang Ma, Xuhong Wang, Yuan Xie, and Xin
  Tan. 2026.
\newblock \href {https://arxiv.org/abs/2601.10744} {Explore with long-term
  memory: A benchmark and multimodal llm-based reinforcement learning framework
  for embodied exploration}.

\bibitem[{Wang et~al.(2025)Wang, Liu, Chen, Zhou, Gan, Zeng, Che, Yu, Hao,
  Shao, Wang, Wu, Wang, Tang, and Hao}]{wang2025guiagentsfoundationmodels}
Shuai Wang, Weiwen Liu, Jingxuan Chen, Yuqi Zhou, Weinan Gan, Xingshan Zeng,
  Yuhan Che, Shuai Yu, Xinlong Hao, Kun Shao, Bin Wang, Chuhan Wu, Yasheng
  Wang, Ruiming Tang, and Jianye Hao. 2025.
\newblock \href {https://arxiv.org/abs/2411.04890} {Gui agents with foundation
  models: A comprehensive survey}.
\newblock \emph{Preprint}, arXiv:2411.04890.

\bibitem[{Wu et~al.(2025)Wu, Wu, Xu, Wang, Sun, Jia, Cheng, Ding, Chen, Liang,
  and Qiao}]{wu2025osatlas}
Zhiyong Wu, Zhenyu Wu, Fangzhi Xu, Yian Wang, Qiushi Sun, Chengyou Jia, Kanzhi
  Cheng, Zichen Ding, Liheng Chen, Paul~Pu Liang, and Yu~Qiao. 2025.
\newblock \href {https://openreview.net/forum?id=n9PDaFNi8t} {{OS}-{ATLAS}:
  Foundation action model for generalist {GUI} agents}.
\newblock In \emph{The Thirteenth International Conference on Learning
  Representations}.

\bibitem[{Xu et~al.(2025)Xu, Wang, Wang, Lu, Xie, Saha, Sahoo, Yu, and
  Xiong}]{xu2025aguvisunifiedpurevision}
Yiheng Xu, Zekun Wang, Junli Wang, Dunjie Lu, Tianbao Xie, Amrita Saha, Doyen
  Sahoo, Tao Yu, and Caiming Xiong. 2025.
\newblock Aguvis: Unified pure vision agents for autonomous gui interaction.
\newblock In \emph{International Conference on Machine Learning}, pages
  69772--69805. PMLR.

\bibitem[{Yadav et~al.(2023)Yadav, Ramrakhya, Ramakrishnan, Gervet, Turner,
  Gokaslan, Maestre, Chang, Batra, Savva
  et~al.}]{yadav2023habitatmatterport3dsemanticsdataset}
Karmesh Yadav, Ram Ramrakhya, Santhosh~Kumar Ramakrishnan, Theo Gervet, John
  Turner, Aaron Gokaslan, Noah Maestre, Angel~Xuan Chang, Dhruv Batra, Manolis
  Savva, and 1 others. 2023.
\newblock Habitat-matterport 3d semantics dataset.
\newblock In \emph{Proceedings of the IEEE/CVF Conference on Computer Vision
  and Pattern Recognition}, pages 4927--4936.

\bibitem[{Yang et~al.(2025{\natexlab{a}})Yang, Yang, Gupta, Han, Fei-Fei, and
  Xie}]{yang2025thinkingspacemultimodallarge}
Jihan Yang, Shusheng Yang, Anjali~W Gupta, Rilyn Han, Li~Fei-Fei, and Saining
  Xie. 2025{\natexlab{a}}.
\newblock Thinking in space: How multimodal large language models see,
  remember, and recall spaces.
\newblock In \emph{Proceedings of the Computer Vision and Pattern Recognition
  Conference}, pages 10632--10643.

\bibitem[{Yang et~al.(2026)Yang, Zeng, Zhong, Jing, Zheng, Chen, Qiu, Qin, Ma,
  and Li}]{yang2025omniactorgeneralistguiembodied}
Longrong Yang, Zhixiong Zeng, Yufeng Zhong, Huang Jing, Liming Zheng, Lei Chen,
  Haibo Qiu, Zequn Qin, Lin Ma, and Xi~Li. 2026.
\newblock \href {https://openreview.net/forum?id=oJAIjUDxkZ} {Omniactor: A
  generalist {GUI} and embodied agent for 2d\&3d worlds}.
\newblock In \emph{The Fourteenth International Conference on Learning
  Representations}.

\bibitem[{Yang et~al.(2025{\natexlab{b}})Yang, Chen, Zhang, Zhao, Qian, Wang,
  Wang, Koripella, Movahedi, Li
  et~al.}]{yang2025embodiedbenchcomprehensivebenchmarkingmultimodal}
Rui Yang, Hanyang Chen, Junyu Zhang, Mark Zhao, Cheng Qian, Kangrui Wang,
  Qineng Wang, Teja~Venkat Koripella, Marziyeh Movahedi, Manling Li, and 1
  others. 2025{\natexlab{b}}.
\newblock Embodiedbench: Comprehensive benchmarking multi-modal large language
  models for vision-driven embodied agents.
\newblock In \emph{International Conference on Machine Learning}, pages
  70576--70631. PMLR.

\bibitem[{Yuan et~al.(2025)Yuan, Duan, Blukis, Pumacay, Krishna, Murali,
  Mousavian, and Fox}]{yuan2024robopointvisionlanguagemodelspatial}
Wentao Yuan, Jiafei Duan, Valts Blukis, Wilbert Pumacay, Ranjay Krishna,
  Adithyavairavan Murali, Arsalan Mousavian, and Dieter Fox. 2025.
\newblock Robopoint: A vision-language model for spatial affordance prediction
  in robotics.
\newblock In \emph{Conference on Robot Learning}, pages 4005--4020. PMLR.

\bibitem[{Zhang et~al.(2024{\natexlab{a}})Zhang, Wu, Yihua, Liao, Xu, Xiao,
  Wei, and Tang}]{zhang2024androidzoochainofactionthoughtgui}
Jiwen Zhang, Jihao Wu, Teng Yihua, Minghui Liao, Nuo Xu, Xiao Xiao, Zhongyu
  Wei, and Duyu Tang. 2024{\natexlab{a}}.
\newblock Android in the zoo: Chain-of-action-thought for gui agents.
\newblock In \emph{Findings of the Association for Computational Linguistics:
  EMNLP 2024}, pages 12016--12031.

\bibitem[{Zhang et~al.(2026)Zhang, Xu, Gong, Wang, Xie, and
  Tan}]{zhang2026worldminecraft}
Lechao Zhang, Haoran Xu, Jingyu Gong, Xuhong Wang, Yuan Xie, and Xin Tan. 2026.
\newblock \href {https://openreview.net/forum?id=dc90uPqxWF} {World2minecraft:
  Occupancy-driven simulated scenes construction}.
\newblock In \emph{The Fourteenth International Conference on Learning
  Representations}.

\bibitem[{Zhang et~al.(2024{\natexlab{b}})Zhang, Wang, Jia, Zheng, Yan, Gao,
  Li, and Xu}]{zhang2024llamatouchfaithfulscalabletestbed}
Li~Zhang, Shihe Wang, Xianqing Jia, Zhihan Zheng, Yunhe Yan, Longxi Gao,
  Yuanchun Li, and Mengwei Xu. 2024{\natexlab{b}}.
\newblock Llamatouch: A faithful and scalable testbed for mobile ui task
  automation.
\newblock In \emph{Proceedings of the 37th Annual ACM Symposium on User
  Interface Software and Technology}, pages 1--13.

\bibitem[{Zheng et~al.(2025{\natexlab{a}})Zheng, Zhou, Bartoldson, Kailkhura,
  Lai, Zhao, and Chen}]{zheng2025actpaysefficientreinforcement}
Haizhong Zheng, Yang Zhou, Brian~R Bartoldson, Bhavya Kailkhura, Fan Lai,
  Jiawei Zhao, and Beidi Chen. 2025{\natexlab{a}}.
\newblock Act only when it pays: Efficient reinforcement learning for llm
  reasoning via selective rollouts.
\newblock In \emph{The Thirty-ninth Annual Conference on Neural Information
  Processing Systems}.

\bibitem[{Zheng et~al.(2025{\natexlab{b}})Zheng, Lu, Wang, Feng, Kuang, and
  Xiong}]{zheng2025easyr1}
Yaowei Zheng, Junting Lu, Shenzhi Wang, Zhangchi Feng, Dongdong Kuang, and
  Yuwen Xiong. 2025{\natexlab{b}}.
\newblock Easyr1: An efficient, scalable, multi-modality rl training framework.
\newblock \url{https://github.com/hiyouga/EasyR1}.

\bibitem[{Zhou et~al.(2025)Zhou, An, Chi, Han, Rong, Zhang, Wang, Wang, Huang,
  Sheng et~al.}]{zhou2025roboreferspatialreferringreasoning}
Enshen Zhou, Jingkun An, Cheng Chi, Yi~Han, Shanyu Rong, Chi Zhang, Pengwei
  Wang, Zhongyuan Wang, Tiejun Huang, Lu~Sheng, and 1 others. 2025.
\newblock Roborefer: Towards spatial referring with reasoning in
  vision-language models for robotics.
\newblock In \emph{The Thirty-ninth Annual Conference on Neural Information
  Processing Systems}.

\end{thebibliography}
\newpage
\appendix

\section{Prompts for reasoning thought generation} \label{Prompts for reasoning thought generation}
Here are our prompts for generating reasoning thoughts.
\begin{tcolorbox}[title=Embodied Thought Generation Prompt,width=\textwidth,breakable,width=\columnwidth,fonttitle=\bfseries]
\footnotesize
You are a robot in an unfamiliar environment. Now I want you to give the reason for your action.

Your action can be in the following list:

\begin{itemize}
    \item Based on the image, predict the optimal location to move next to finish the task. Use the coordinates (x, y) (x is the pixel from left to right and y is the pixel from top to bottom) to indicate where you want to move to: \\
    \texttt{\{"action\_type": "move", "x": <position in horizontal (width)>, "y": <position in vertical (height)>\}}.
    
    \item Turn left: \texttt{\{"action\_type": "turn\_left"\}}.
    
    \item Turn right: \texttt{\{"action\_type": "turn\_right"\}}.
    
    \item Turn around: \texttt{\{"action\_type": "turn\_around"\}}.
    
    \item Move the camera angle downward: \texttt{\{"action\_type": "look\_down"\}}.
    
    \item Based on the image, if you find the target and the target is close enough, please stop to indicate that you want to stop: \\
    \texttt{\{"action\_type": "stop"\}}.
\end{itemize}

You will be given the view before you performed the action (which has a text label \texttt{"before"} on the bottom right), the action you chose, and the task.

\textbf{This is the action you performed:} \texttt{<\textit{action}>} \\
\textbf{This is the task:} \texttt{<\textit{task/question}>} (the picture and action is one of the steps to finish the task)

By inspecting the picture and the action performed, give a brief reason of this step. You should carefully inspect the environment and give your analysis for why to do such action rather than other actions. 

If moving to a position, explain why moving to that position based on the current environment. Avoid generic reasons like “get closer to the target.”

\textbf{NOTICES:}
\begin{enumerate}
    \item Coordinates are absolute coordinates (a center point defined by top-left and bottom-right coordinates).
    \item If the action type is \texttt{"move"}, the point will be labeled as \texttt{"Next point"} in the before image.
    \item Remember that you should give the answer from a first-person perspective and keep it around 60 words and in a single line.
    \item Don’t limit yourself to begin with “I...”. try any other possible sentence structure(like the position of exchangeing description and target) if not influence the meaning."
\end{enumerate}
\end{tcolorbox}
\begin{tcolorbox}[title=GUI Thought Generation Prompt,width=\textwidth,breakable,width=\columnwidth,fonttitle=\bfseries]
\footnotesize
You are an agent who can operate an Android phone on behalf of a user.  
Now I want you to give the reason for your action.

You will be given the screenshot before you performed the action (which has a text label "before" on the bottom right),  
the action you chose (together with the reason), and the screenshot after the action was performed (which has a text label "after" on the bottom right).

\textbf{This is the action you picked:} \texttt{<\textit{q\_text}>}

\textbf{This is the task:} \texttt{<\textit{task>}}  
(The screenshots and action are one of the steps to finish the task)

\textbf{This is the instruction:} \texttt{<\textit{instruction>}}  
(The instruction to solve the task)

\textbf{This is the related apps:} \texttt{<\textit{apps>}}  
(Apps in the reason you output cannot go beyond the range of the app list)

By comparing the two screenshots and the action performed, give a brief reason of this step.  
The reason should include the detailed description for the action and the target to do so, but avoid any description related to the after screenshot.

\textbf{Requirements:}
\begin{itemize}
    \item Use first-person perspective.
    \item Keep the response around 60 words and in a single line.
    \item Do not begin every sentence with "I"; feel free to vary the structure as long as the meaning remains clear.
\end{itemize}
\end{tcolorbox}
\section{Prompts for training} \label{Prompts for training}
Here is our prompts for training NaviMaster.

\begin{tcolorbox}[title=Spatial-referring  Prompt,breakable,width=\columnwidth]
\footnotesize
Your answer should be formatted as a tuple, i.e. \verb|[x, y]|, where the tuple contains the x and y coordinates of a point satisfying the conditions above.

Output the thinking process in \texttt{<think> ... </think>} tags, and the final answer in:

\begin{raggedright}
\texttt{<answer>[{"action": "moveto", "point": [x, y]}]</answer>}
\end{raggedright}

Note: The coordinates should be between the size of picture, indicating the absolute pixel locations of the points in the image.

Example: \\
\texttt{[{"action": "moveto", "point": [123, 300]}]}
\end{tcolorbox}

\begin{tcolorbox}[title=Navigation Prompt,breakable,width=\columnwidth,fonttitle=\bfseries]
\footnotesize
"You are a Navigation Robot in an unfamiliar environment. In this photo \textless image\textgreater, the task is '\{text\}', with the history being '\{history\}'\\
You need to use your prior knowledge about where items are typically located within a home.\\
Please predict next action to find target item.\\
Your action can be in the following list:\\
Basic Action(move to a point on the ground in the picture):\\
- Based on the image, predict the optimal location to move next to finish the task, use the coordinates (x, y)(x is the pixel from left to right and y is the pixel from top to bottom) to indicate where you want to move to:\\
\texttt{[\{"action": "moveto", "point": [x(position in horizontal(width)), y(position in vertical(height))]\}].}\\
View Adjustment Actions(adjust view as current photo does not have suitable position):\\
- Executes a 90-degree rotation to the left from the current facing direction. Ideal for navigating around obstacles on the right, aligning with a leftward path, or adjusting the view to inspect the left side of the environment. Use this when the task requires a lateral shift to the left: \\
\texttt{[\{"action": "turn\_left"\}].}\\
- Rotates the perspective 90 degrees to the right. This action is useful when the target object or destination is positioned on the right, or when you need to change the direction to follow a rightward route: \\
\texttt{[\{"action": "turn\_right"\}].}\\
- Performs a 180-degree rotation, flipping the orientation to face the opposite direction. This is valuable for finding a possible way if there is no path in front of you: \\
\texttt{[\{"action": "turn\_around"\}].}\\
- Adjusts the camera view to look downwards, without physically moving the position. This is particularly useful for examining details on the ground, such as identifying objects, reading markings, or inspecting lower-level structures: 
\texttt{[\{"action": "look\_down"\}].}\\
Stop Action:\\
- Carefully inspect the environment and judge from history, if you find your target in your view and has been close enough for about 1 meter, stop at current position: \\
\texttt{[\{"action": "stop"\}].}\\
Output the thinking process in \textless think\textgreater \textless/think\textgreater\ tags, and the final answer in \textless answer\textgreater \textless/answer\textgreater\ tags as follows:\\
\textless think\textgreater ... \textless/think\textgreater\ \textless answer\textgreater answer here \textless/answer\textgreater\\
Note: The 'point' should contain the coordinates of the next destination. Coordinates are absolute coordinates(a center point defined by top-left and bottom-right coordinates). Ensure the predicted Example:\\
\texttt{\{"action": "moveto", "point": [123, 300]\}}\\
\end{tcolorbox}

\begin{tcolorbox}[title=GUI Prompt,width=\textwidth,breakable,width=\columnwidth,fonttitle=\bfseries]
\footnotesize
A conversation between User and Assistant. The user asks a question, and the Assistant solves it step by step. The assistant first thinks about the reasoning process in the mind and then provides the user with the answer. 

At each step, you will be given the current screenshot and the history of the conversation (include screenshot and action in each step). Based on these pieces of information and the goal, you must give the whole content of what you think and then choose to perform one of the actions in the following list (action description followed by the JSON format) by outputting the action in the correct JSON format.
Click/tap on an element on the screen. We have defined the width and height of the screenshot, use the coordinates (x, y) (x is the pixel from left to right and y is the pixel from top to bottom) to indicate which element you want to click, both x and y are integers:\\
\texttt{[\{"action": "click", "point": [x(position in horizontal(width)), y(position in vertical(height))]\}]}\\
Long press on an element on the screen, similar with the click action above, use the coordinates (x, y) to indicate which element you want to long press:\\  
\texttt{[\{"action": "long\_press", "point": [x(position in horizontal(width)), y(position in vertical(height))]\}]}\\
Type text into a text field (this action contains clicking on the target field, typing in the text and pressing the enter), use the coordinates (x, y) to indicate which element you want to click, both x and y are integers:\\
 \texttt{[\{"action": "input\_text", "text": <text\_input>, "point": [x(position in horizontal(width)), y(position in vertical(height))]\}]}\\
 Navigate to the home screen:\\
 \texttt{[\{"action": "navigate\_home"\}]}\\
Navigate back:\\
\texttt{[\{"action": "navigate\_back"\}]}\\
Scroll the screen or a scrollable UI element from start point to end point, use the coordinates (x, y) to indicate the two points you want to scroll:\\
\texttt{[\{"action": "scroll", "start\_point": [<start position in horizontal(width)>, <start position in vertical(height)>], "end\_point": [<end position in horizontal(width)>, <end position in vertical(height)>]\}]}\\
NOTICES:
1.Coordinates are absolute coordinates (a center point defined by top-left and bottom-right coordinates).
2.The reasoning process and answer are enclosed within \verb|<think>| \ldots \verb|</think>| and \verb|<answer>| \ldots \verb|</answer>| tags, respectively.
Example: \\
\texttt{<think> reasoning process here </think>
<answer>[{"action": "click",
"point": [378, 871]}]</answer>}

\end{tcolorbox}

\section{Action Space} \label{action space}
In our trajectory, the action space $\mathcal{A}_{gui}$ for GUI task is defined as:
\begin{equation}
\mathcal{A}_{gui} = 
\begin{cases}
\textbf{CLICK} \ (x, y)\\
\textbf{SCROLL} \ [up, down, right, left]\\
\textbf{LONGPRESS} \ (x, y) \\
\textbf{TYPE} \ [TEXT] \ (x, y) \\
\textbf{HOME} \\
\textbf{BACK} \\
\end{cases} \ 
\end{equation}
The action space $A_{emb}$ for embodied task is defined as:
\begin{equation}
\mathcal{A}_{emb} = 
\begin{cases}
\textbf{MOVETO} \ (x, y)\\
\textbf{TURN} [left, right, around, down] \\
\textbf{STOP} \\
\end{cases} \ 
\end{equation}

\section{Training Settings} \label{Hyperparameter}
\subsection{Training Hyperparameter}
To ensure the fairness of all comparative and ablation experiments, we maintained consistent hyperparameter settings throughout the training process, as detailed in Table \ref{appendix: Hyperparameter}.
\begin{table}[htbp]
\centering
\begin{tabular}{@{}ll@{}}
\toprule
Hyperparameter & Value \\ \midrule
learning\_rate & from 1e-6 to 0 \\
temperature & 1.0 \\
num\_generations & 5 \\
num\_train\_epochs & 3 \\
max\_prompt\_length & 7000 \\
max\_response\_length & 1024 \\
per\_device\_train\_batch\_size & 4 \\
gradient\_accumulation\_steps & 16 \\
KL coefficient & 0.01 \\ 
Reward coefficients $\lambda_1$, $\lambda_2$, $\lambda_3$  & 0.1, 1, 1\\
\bottomrule
\end{tabular}
\caption{Hyperparameter settings used for all reinforcement learning training.}
\label{appendix: Hyperparameter}
\end{table}

In our implementation, whenever the concatenated trajectory exceeds 7000 tokens, we iteratively remove the minimum number of oldest steps until the prompt length drops just below the 7000 limit.

\subsection{Sampling strategy}
\begin{figure}[t]
    \centering
    \includegraphics[width=\columnwidth]{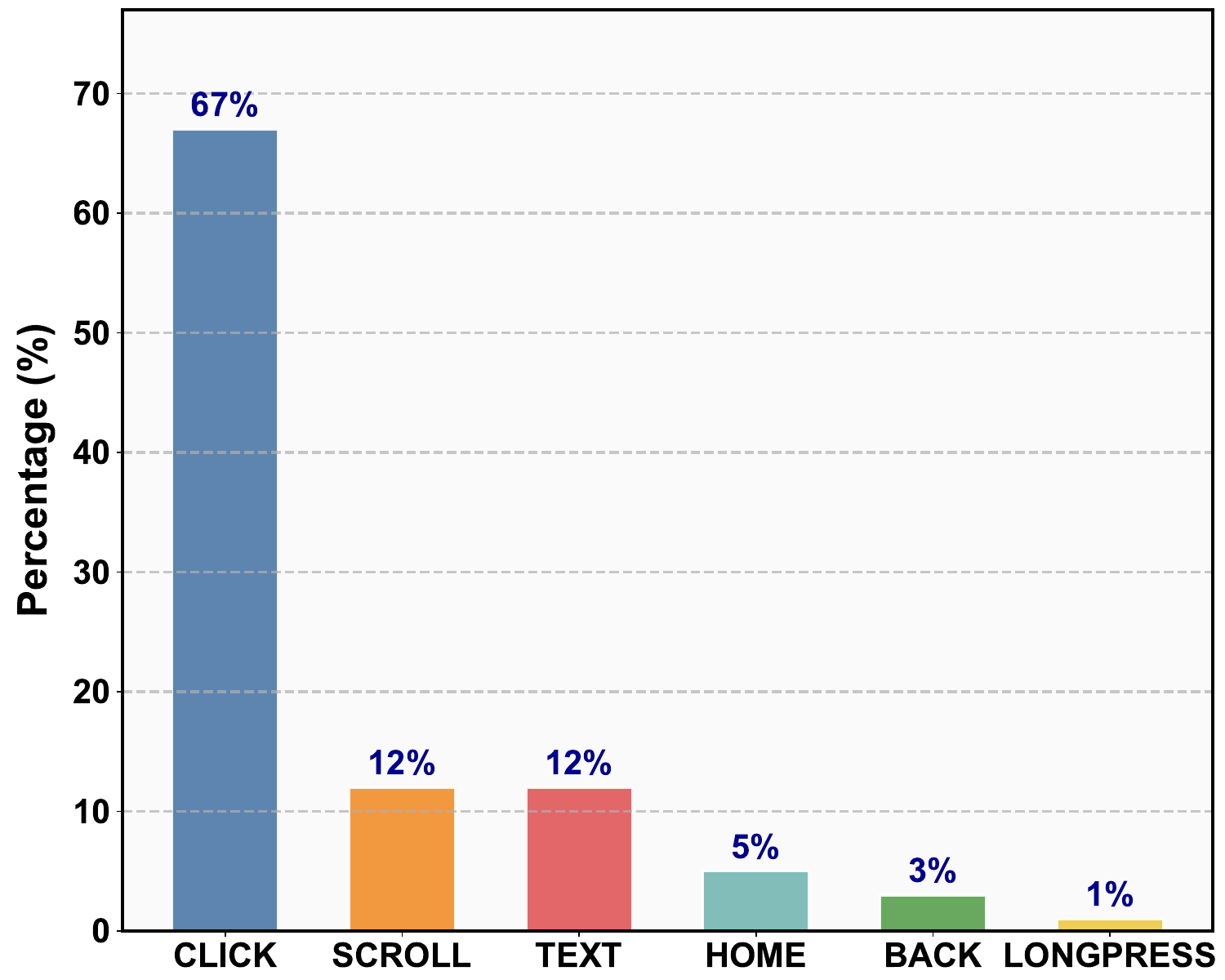}
    \caption{Data distribution of GUI-Odyssey}
    \label{fig:action_distribution}
\end{figure}

Selection bias could significantly affect the results. To avoid this, our sampling strictly follows the original distribution of GUI-Odyssey. We first computed the action-type distribution in the full 173k dataset. As shown in Fig.~\ref{fig:action_distribution}, when sampling the subset, we preserved this distribution, ensuring that the sampled data statistically matches the source dataset. Therefore, no additional bias was introduced during the sampling process.

\section{GUI Benchmark and Metrics Details} \label{metrics details}
All the GUI benchmark are the test set of open-source dataset. The in-domain testing is from our GUI data source. We sample 4800 cases to test our in-domain performance.

In GUI task, we follow the settings in OS-Atlas, where a correct type prediction is considered accurate if the predicted action type matches the ground truth. For predictions involving grounding, an action is deemed correct if the predicted location falls within 14\% of the image size relative to the ground truth.

Here, we argue that the metric of Type (the accuracy of the predicted action type) is unreliable due to dataset bias. As shown in Table~\ref{inbalance}, a naive model that predicts all actions as [\textbf{CLICK}] can still achieve a high Type prediction accuracy. Despite this limitation, we report Type results for completeness in Table \ref{tab:type_performance}.

\begin{table}[htbp]
\centering	
\tiny
\renewcommand{\arraystretch}{0.65}
\begin{tabular}{c@{\hspace{3pt}}c@{\hspace{3pt}}c@{\hspace{3pt}}c@{\hspace{3pt}}c@{\hspace{3pt}}c}
\toprule
  Metric & AC-High/Low &AITW &GUIAct-Phone&LlamaTouch&AITZ\\
    \midrule
   Type& 59.7&57.2 &58.0&64.4&55.9\\
    \hline
\end{tabular}
 \vspace{-2mm}
\caption{The bias of action types in testing datasets.} 
\label{inbalance}
\end{table}

\renewcommand{\arraystretch}{0.7}
\begin{table*}[t]
    \centering
    \tiny
    \begin{tabular}{lcccccccccc}
        \toprule
        & \textbf{AC-Low} & \textbf{AC-High} & \textbf{AITW} & \textbf{GuiAct-P} & \textbf{Llamatouch} & \textbf{AITZ} & \textbf{GuiAct-W} & \textbf{OmniAct-W} & \textbf{OmniAct-D} & \textbf{Odyssey} \\
        \midrule
        Ours (w/o Embodied) & 81.16 & 69.50 & 71.47 & 71.10 & 88.21 & 64.25 & 91.21 & 95.25 & 95.79 & 78.54 \\
        Ours (w/o GUI) & 80.47 & 71.45 & 53.15 & 51.44 & 90.15 & 45.29 & 83.62 & 93.82 & 90.15 & 71.30 \\
        Ours & 81.08 & 73.88 & 67.47 & 68.71 & 86.38 & 61.79 & 92.13 & 94.06 & 95.62 & 77.51 \\
        \bottomrule
    \end{tabular}
    \caption{TYPE results on different GUI tasks.}
    \label{tab:type_performance}
\end{table*}

\section{Embodied Benchmark and Metrics Details} \label{embodied metrics details}
For details of benchmark:

\textbf{Where2Place.} This benchmark contains 100 real-world images to evaluate free space referring.

\textbf{RoboSpatial.} There are three branch in the benchmark:``Configuration'', ``Context'' and ``Compatibility''. We take the ``Context'' branch to test free space referring. 

\textbf{RefSpatial.} We take the unseen set of RoboSpatial. This set comprises 77 samples from the Location/Placement task.

\textbf{Roboreflt.} We take the testA set of Roboreflt.

For metrics, we introduce the average success rate of predicted points with in the groundtruth mask to evaluate the spatial grounding accuracy in the spatial referring task. This metric directly assesses the model's ability to accurately localize the target based on the natural language description. For the navigation task, consistent with prior works, we utilize Success Rate (SR) and Success Rate Weighted by Inverse Path Length (SPL) as our metrics. SR measures the percentage of episodes that are successfully completed. Here, we set the success threshold to 0.3, meaning that stopping within this distance from the goal will be considered a success. SPL is a measure of navigation path efficiency, which quantifies the agent's performance by considering both task success and the path efficiency relative to the optimal path.
\section{Detailed Analysis}
\label{Detailed Analysis}

\subsection{Statistical Stability and Error Statement}
Our results are based on multiple runs and are empirically stable. All reported results are averaged over multiple trials, and to reduce variance during inference, we set the vLLM~\cite{kwon2023efficient} temperature to 0. Moreover, the success criteria for our tasks are range-based rather than relying on a single exact output. Therefore, even if model predictions vary slightly due to stochasticity, the evaluation outcome is generally unaffected.

\subsection{Detailed statistics}
The detailed results of base model ablation, data scales ablation, data scale ablation and reward design ablation is shown in the Table.\ref{Detailed statistics}. 
\begin{table}[htbp]
\centering	
\tiny
\renewcommand{\arraystretch}{0.85} 
\begin{tabular}{@{}l@{\hspace{3pt}}c@{\hspace{3pt}}c@{\hspace{3pt}}c@{\hspace{3pt}}c@{}}
\toprule
     \bf{Models} &\bf{AC-High}&\bf{AC-Low} &\bf{Where2Place}&\bf{RefSpatial}\\
    \midrule
    Qwen2.5VL-3B &64.57/34.43 &85.58/58.78&2.96&3.36 \\
    Navimaster-3B &66.02/45.26 &79.85/60.41 &14.83&15.59\\
    Navimaster-3B(w/o GUI) & 45.89/28.75 & 75.21/56.25 &7.00&7.79\\
    Navimaster-3B(w/o Embodied) &45.89/28.75 &75.21/56.25&7.00&7.79 \\
    Qwen2VL-7B &19.11/2.30 &17.98/7.24&3.00&1.30 \\
    Navimaster-qwen2vl &24.18/14.43 &31.54/18.81 &22.95&7.79\\
    Navimaster-qwen2vl(w/o GUI) & 19.37/13.04 & 26.46/15.71 &15.98&9.09\\
    Navimaster-qwen2vl(w/o Embodied) &23.53/13.87 &30.65/18.74&15.01&1.30 \\
    Navimaster-7k &76.90/52.66&93.85/70.41&41.07&18.19 \\
    Navimaster-7k(w/o GUI) &26.86/29.06 &77.78/63.37&44.94&15.08 \\
    Navimaster-7k(w/o Embodied) &74.83/52.34 &94.13/69.48&37.73&14.28 \\
    Navimaster-1gui2embodied &74.87/54.83 &94.19/70.09&47.77&18.82 \\
    Navimaster-1embodied2gui &74.55/54.73 &93.64/71.04&47.97&19.21 \\
    Navimaster-hardreward &76.19/54.07 &92.56/68.39&44.01&14.28 \\
    Navimaster-robopoint &73.32/52.34 &93.94/68.94&52.04&19.67 \\
    Navimaster-navi &73.64/52.44 &93.90/69.37&47.99&16.49 \\
    Navimaster-1\_1\_1 &75.03/53.20 &94.20/69.31&49.96&20.14 \\
    Navimaster-0.1\_1\_2 &77.33/54.98 &93.70/69.19&47.06&22.14 \\
    Navimaster-0.1\_2\_1 &76.04/54.10 &94.06/70.72&46.06&23.34 \\
    \hline
\end{tabular}
\caption{Detailed Statistics} 
\label{Detailed statistics}
\end{table}
\subsection{Embodied Data Source} 
Our embodied data consists of two parts: one part is derived from the point-based data we construct (trajectory), and the other part is the spatial affordance prediction data from RoboPoint (affordance). We evaluate the model when trained on each source individually and on their union (affordance + trajectory). As shown in the Table.\ref{Detailed statistics}, Navimaster-robopoint the embodied data source only from affordance data. Navimaster-navi represents model  embodied data are only from trajectory. Results show that combining both sources under an equal total data volume yields the best training performance.
\subsection{Data Usage} 
\begin{figure}[t]
    \includegraphics[width=\columnwidth]{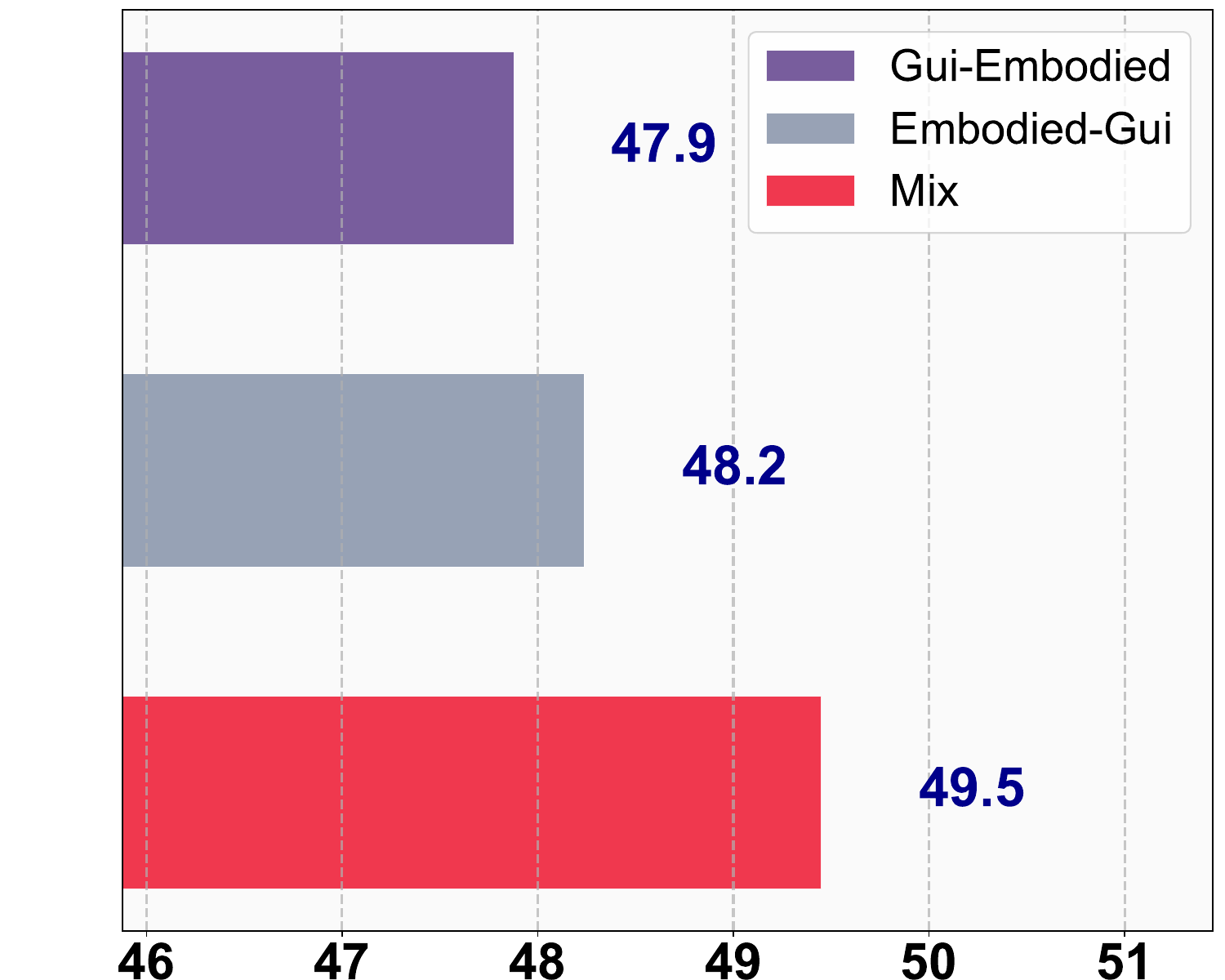}
    \caption{Performance of different data usage strategies.}
    \label{fig:data usage strategy}
\end{figure}

Regarding dataset utilization, there are primarily two strategies. One is to mix different types of data into a single training phase (Mix). The other is to adopt a multi-stage schedule, with each stage focusing on one specific task or subset of data (GUI-Embodied or Embodied-GUI). Details were shown in the Table.\ref{Detailed statistics}, Navimaster-1gui2embodied means training on GUI data at stage 1 and then on embodied data for stage 2. Navimaster-1embodied2gui represents model trained on embodied data first and then on GUI data. The average results shown in Fig.~\ref{fig:data usage strategy}, the mix training strategy generally outperforms the two-stage training strategy across various benchmarks. This suggests that training with mix data in a single phase enables the model to exploit complementary information effectively.
\subsection{Hyperparameters}
The ablation on hyperparameters focuses on the three reward weights, $\lambda_1$, $\lambda_2$, and $\lambda_3$. We conducted three additional experiments with configurations different from the main setup:
NaviMaster\_0.1\_1\_2 ($\lambda_1=0.1$, $\lambda_2=1$, $\lambda_3=2$),
NaviMaster\_0.1\_2\_1 ($\lambda_1=0.1$, $\lambda_2=2$, $\lambda_3=1$), and
NaviMaster\_1\_1\_1 ($\lambda_1=1$, $\lambda_2=1$, $\lambda_3=1$).
As shown in Table~\ref{Detailed statistics}, the configuration used in our main experiments achieves the best overall performance among these variants.

\subsection{Reward Components}
We attribute the performance improvements to gains in both navigation accuracy and instruction adherence. To decouple these factors, we provide the reward components ablation study in Table~\ref{Ablation of Reward Components}. It shows that removing the format or type reward leads to a drop (1–2 points) across all navigation benchmarks.
\begin{table}[htbp]
\centering	
\tiny
\renewcommand{\arraystretch}{0.85} 
\begin{tabular}{l@{\hspace{3pt}}c@{\hspace{3pt}}c@{\hspace{3pt}}c@{\hspace{3pt}}c}
\toprule
     Reward &\bf{AC-High}&\bf{AC-Low} &\bf{Where2Place}&\bf{RefSpatial}\\
    \midrule
 NaviMaster        & 55.89   & 69.46  & 52.97       & 19.49      \\
 w/o format reward & 54.02   & 68.95  & 52.02       & 20.77      \\
 w/o type reward   & 55.66   & 70.60  & 48.99       & 18.18      \\
    \hline
\end{tabular}
\caption{Ablation of Reward Components} 
\label{Ablation of Reward Components}
\end{table}

\section{Discussion of Threshold} \label{Discussion of Threshold}
\subsection{Standardized High Resolution}
 We did not use low-resolution images (e.g., $224 \times 224$ or $480 \times 640$) for embodied observations. Instead, we maintained a high-resolution standard: the minimum GUI resolution is $1280 \times 720$, and all embodied observations are uniformly resized to $1980 \times 1080$. This decision was deliberate, as prior works have shown that using high-resolution images during training and inference significantly enhances the model's crucial grounding capability.
\subsection{Rationale for Absolute Threshold}
We chose an absolute threshold over a relative one to prevent severe reward hacking. If a relative threshold were used, on larger-resolution images, the acceptable error range would scale proportionally, making the success condition much easier. This would lead to many localization-inaccurate rollouts being incorrectly deemed "successful", which would seriously mislead the model's training process.The absolute threshold of 200 pixels avoids this issue, ensuring that only highly accurate samples receive a positive reward, thereby preventing incorrect samples from skewing the training, even if it makes successful rollouts marginally more challenging on high-resolution data.
\subsection{Alignment with Established Metrics}
Furthermore, the $\theta_d = 200$ pixels threshold is not arbitrary; it is consistent with widely accepted evaluation metrics in the GUI community. Success in GUI tasks is commonly judged by whether the predicted action is within a certain area (e.g., $\mathbf{14\%}$) of the ground truth. Given our minimum resolution of $1280 \times 720$, a 200-pixel radius circle aligns almost perfectly with this standard:$$\pi \times (200^2) / (1280 \times 720) \approx 14\%$$
\subsection{Training with relative threshold}
We are prepared to conduct additional experiments using a relative threshold for reward definition. Specifically, we would normalize the model prediction and ground truth to a $0-1000$ range and set the reward threshold to $\mathbf{140}$ for training. The Table~\ref{relative threshold} shows the advantage of an absolute reward threshold.
\begin{table}[htbp]
\centering	
\tiny
\renewcommand{\arraystretch}{0.85} 
\begin{tabular}{l@{\hspace{3pt}}c@{\hspace{3pt}}c@{\hspace{3pt}}c@{\hspace{3pt}}c}
\toprule
     \bf{threshold} &\bf{AC-High}&\bf{AC-Low} &\bf{Where2Place}&\bf{RefSpatial}\\
    \midrule
     absolute   & 55.89    & 69.46   & 52.97 & 19.49 \\
     relative   & 53.61    & 67.56   & 51.99 & 18.48  \\
    \hline
\end{tabular}
\caption{Comparison of relative and absolute threshold} 
\label{relative threshold}
\end{table}
\subsection{Ablation of threshold $\theta_d$}
We provide experiments in Table~\ref{threshold d}, which evaluates different distance thresholds to further validate the robustness of our formulation.
\begin{table}[htbp]
\centering	
\tiny
\renewcommand{\arraystretch}{0.85} 
\begin{tabular}{l@{\hspace{3pt}}c@{\hspace{3pt}}c@{\hspace{3pt}}c@{\hspace{3pt}}c}
\toprule
     $\theta_d$ &\bf{AC-High}&\bf{AC-Low} &\bf{Where2Place}&\bf{RefSpatial}\\
    \midrule
 100 pixel  & 55.06    & 69.85   & 48.88 & 19.48  \\
 200 pixel  & 55.89    & 69.46   & 52.97 & 19.49  \\
 300 pixel  & 55.77    & 69.60   & 48.14 & 16.88  \\
 400 pixel  & 53.61    & 68.58   & 51.06 & 16.68  \\
    \hline
\end{tabular}
\caption{Ablation of threshold $\theta_d$} 
\label{threshold d}
\end{table}

\subsection{Ablation of threshold $\theta_h$}
We clarify that the introduction of depth into the reward signal is primarily intended for robustness, not for a direct enhancement of policy performance, thereby reducing the reliance on sensor priors as the main source of improvement. If the reward were based purely on 2D distance from the image projection, it would fail in corner cases—for example, when an object far away and an object nearby appear spatially aligned and close in the 2D image. Such misleading 2D calculations necessitate the introduction of depth to compute the true 3D distance, ensuring a robust and accurate proximity reward signal. We have conducted an ablation study on the effect of the depth signal on the overall performance in Table~\ref{threshold h}.
\begin{table}[htbp]
\centering	
\tiny
\renewcommand{\arraystretch}{0.85} 
\begin{tabular}{l@{\hspace{3pt}}c@{\hspace{3pt}}c@{\hspace{3pt}}c@{\hspace{3pt}}c}
\toprule
     $\theta_h$ &\bf{AC-High}&\bf{AC-Low} &\bf{Where2Place}&\bf{RefSpatial}\\
    \midrule
 w/o  $\theta_h$ & 55.09    & 69.46   & 52.16 & 19.49    \\
 0.1 m  & 51.56    & 67.98   & 36.84 & 16.88             \\
 1 m  & 55.89    & 69.46   & 52.97 & 19.49             \\
 10 m  & 55.89    & 69.46   & 52.97 & 19.49             \\
    \hline
\end{tabular}
\caption{Ablation of threshold $\theta_h$} 
\label{threshold h}
\end{table}
These results confirm that the presence of the depth signal does not lead to a significant performance boost (it is not a primary source of gain). However, setting an overly strict depth threshold ($\theta_h$) severely limits the reward availability and degrades performance, which supports our claim that depth's role is to ensure robustness by preventing misleading 2D projection errors, not to artificially inflate the success rate.

\section{Case Study}
\begin{figure*}
\centering
   \includegraphics[width=\textwidth]{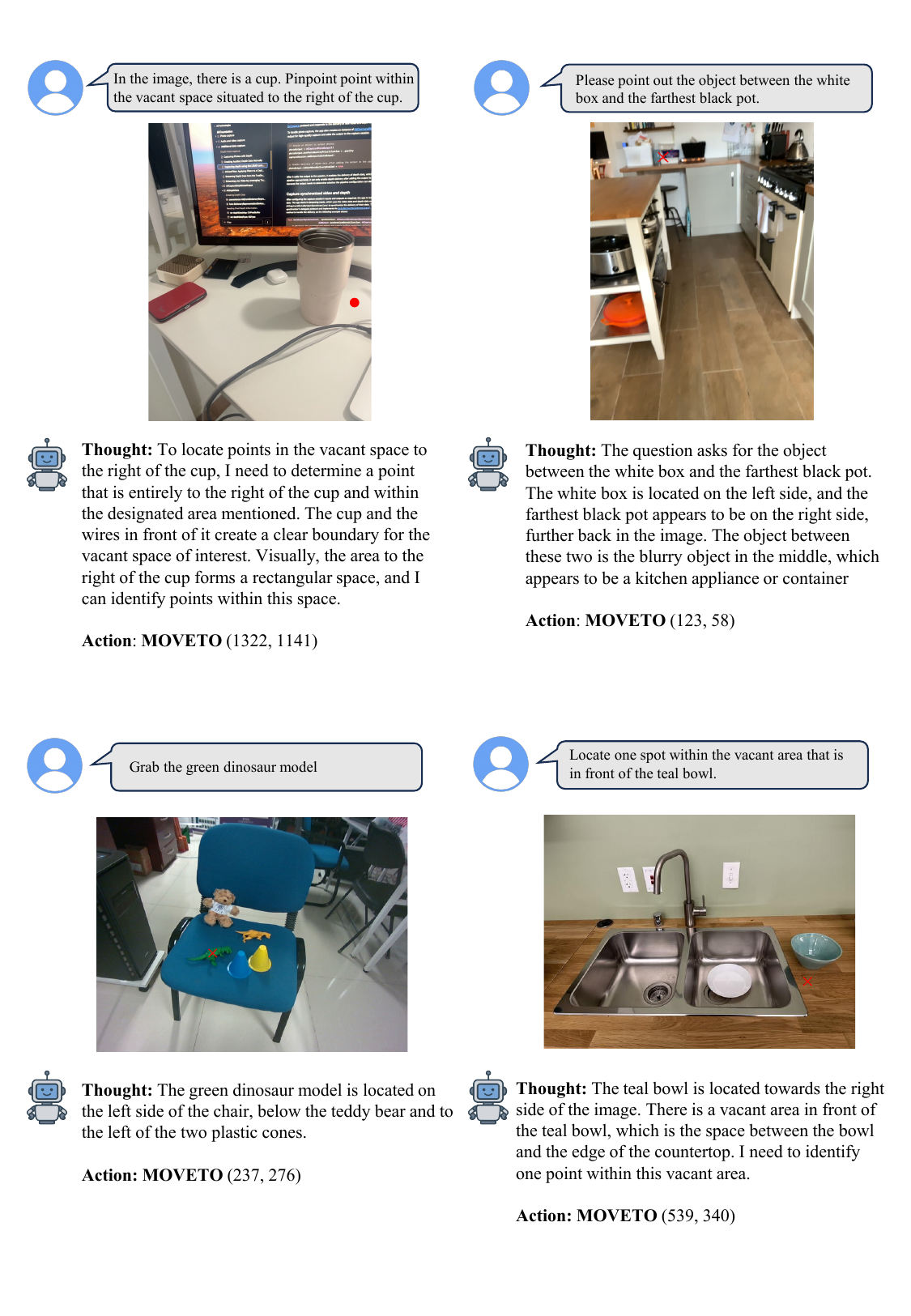}
   \caption{Case studies in RefSpatial, RoboSpatial, RoboReflt and Where2Place.}
   \label{casestudy}
\end{figure*}

Fig.~\ref{casestudy} illustrates the spatial referring task, where the agent must local the coordinates in the image according to the given instruction. Our case set comprises three free space referring instances and one object referring instance.

\begin{figure*}
\centering
   \includegraphics[width=\textwidth]{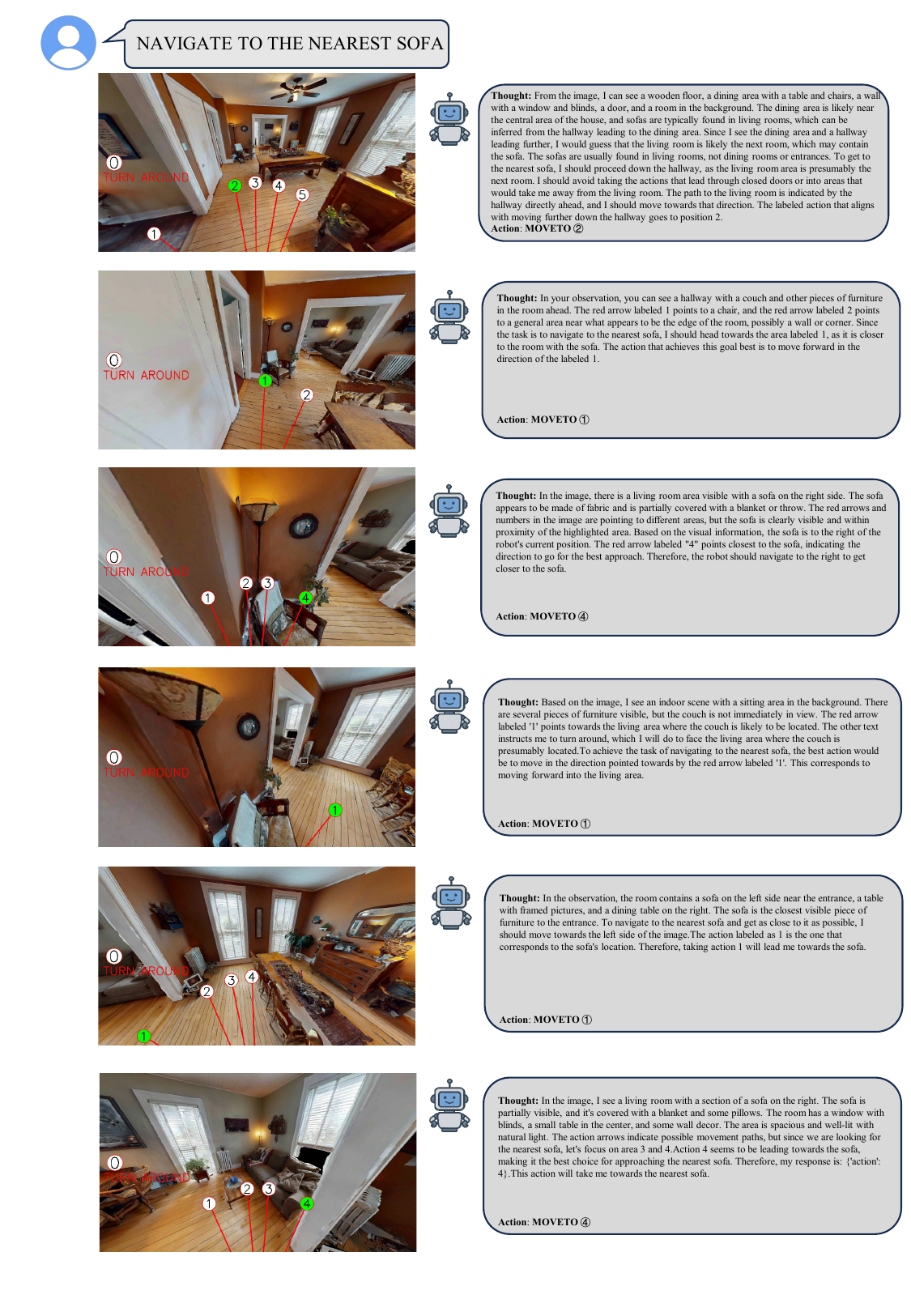}
   \caption{Case study in ObjectNav.}
   \label{casestudy2}
\end{figure*}

Fig.~\ref{casestudy2} illustrates the Embodied Navigation task, where the agent must follow the instruction to complete an entire navigation episode in Habitat.

\begin{figure*}
\centering
   \includegraphics[width=\textwidth]{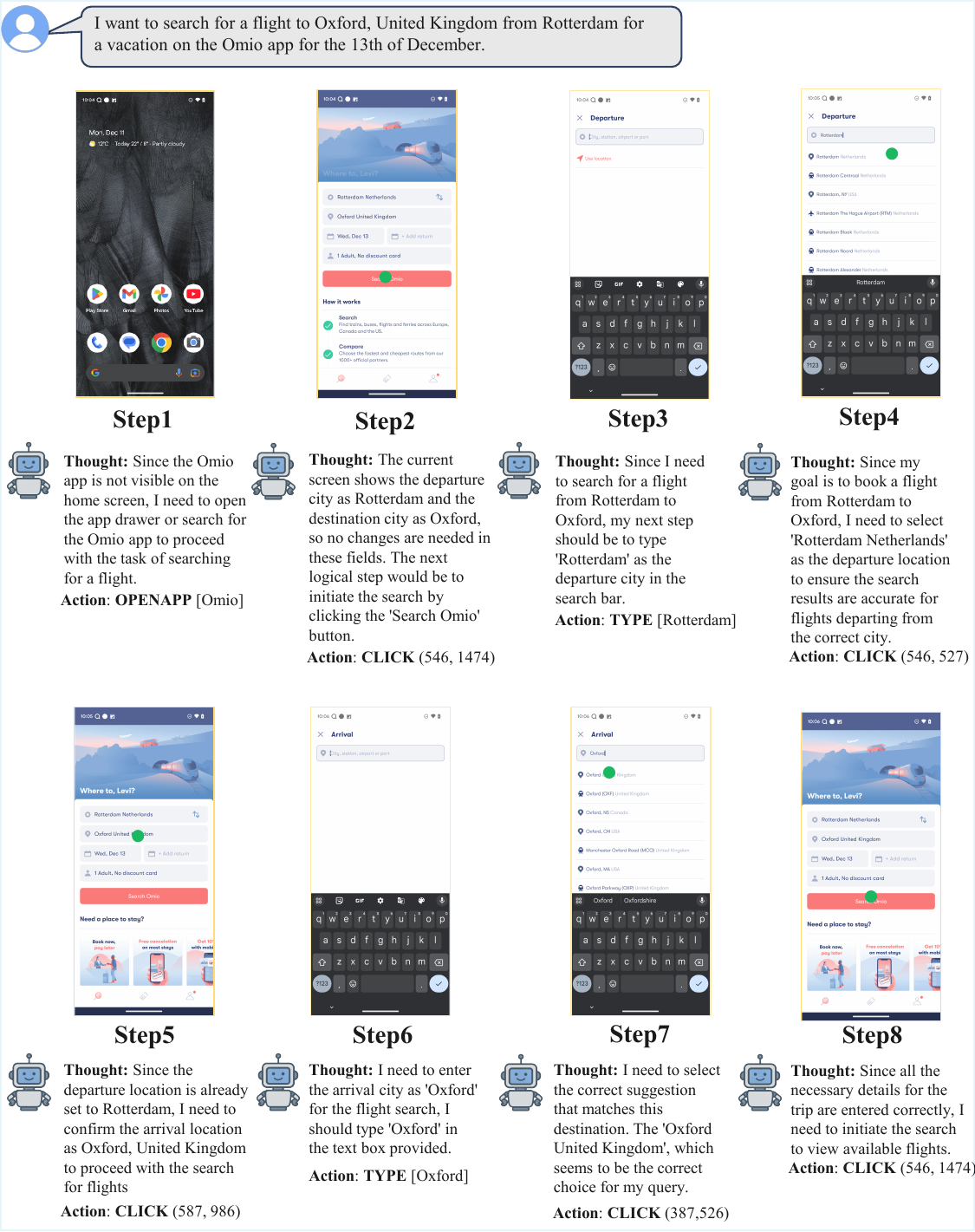}
   \caption{Case study in AC-High.}
   \label{casestudy3}
\end{figure*}

\begin{figure*}
\centering
   \includegraphics[width=\textwidth]{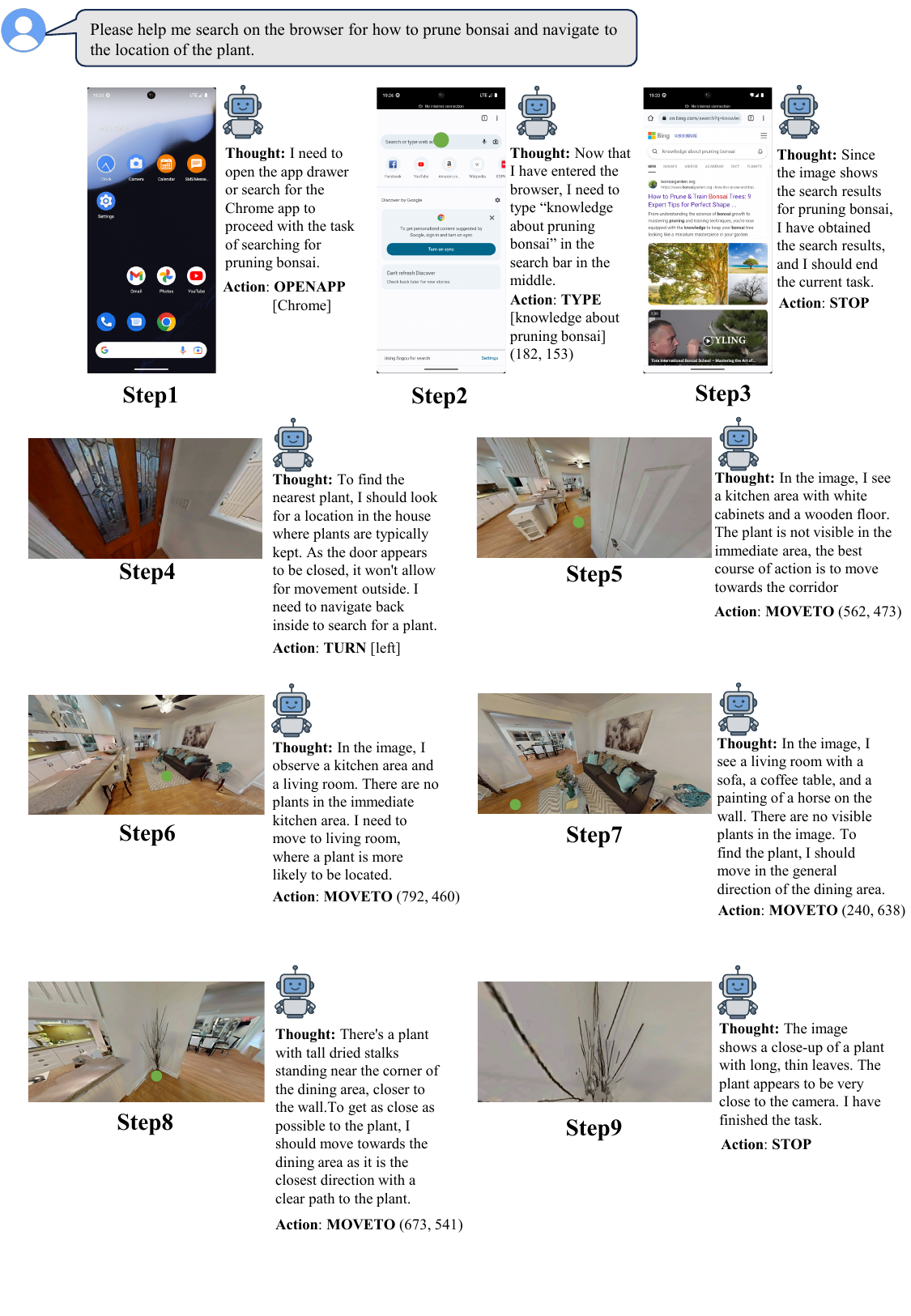}
   \caption{Case study in general task.}
   \label{casestudy4}
\end{figure*}

Fig.~\ref{casestudy3} illustrates the AC-High task, where the agent receives a high-level user instruction and, based on the history and the current smartphone screenshot, outputs the action for the current step.
Fig.~\ref{casestudy4} illustrates the general task, where the agent receives an user instruction and face both GUI and embodied environments. It outputs the thoughts and action for the current step.

\end{document}